%% file: main.tex
\documentclass{article} 
\usepackage{iclr2027_conference,times}

\input{math_commands.tex}

\usepackage[hyphens]{url}
\usepackage{hyperref}
\hypersetup{colorlinks=true,linkcolor=blue!60!black,citecolor=blue!60!black,urlcolor=blue!60!black}
\usepackage{graphicx}
\usepackage{natbib}
\usepackage{caption}
\usepackage{amsmath,amssymb}
\usepackage{multirow}
\usepackage{comment}
\usepackage{booktabs}
\usepackage[table]{xcolor}
\usepackage{adjustbox}
\usepackage{algorithm}
\usepackage{algorithmic}
\graphicspath{{Figures/}{iclr2027/Figures/}{./}}

\IfFileExists{bbm.sty}{\usepackage{bbm}\newcommand{\ind}{\mathbbm{1}}}{\newcommand{\ind}{\mathbf{1}}}

\newif\ifdraftmode
\draftmodefalse
\ifdraftmode
  \newcommand{\todo}[1]{{\color{red}\textbf{[TODO: #1]}}}
\else
  \newcommand{\todo}[1]{}
\fi

\definecolor{darkgreen}{RGB}{0,100,0}
\newcommand{\dn}{\textcolor{red}{$\downarrow$}}
\newcommand{\up}{\textcolor{green!50!black}{$\uparrow$}}
\newcommand{\WR}{WR\textsuperscript{2}}
\newcommand{\yref}{y_{\mathrm{ref}}}

\title{UQ-LOB: Uncertainty-Aware Limit Order Book Mid-Price Forecasting}

\author{Derrick Gilchrist Edward Manoharan$^{*}$ \\
Data Science Research Centre \\
Tampere University \\
Tampere, Finland \\
\texttt{derrick.edwardmanoharan@tuni.fi}
\thanks{Corresponding author.}
\And
Eljas Linna \\
Data Science Research Centre \\
Tampere University \\
Tampere, Finland \\
\texttt{eljas.linna@tuni.fi}
\And
Kestutis Baltakys \\
Data Science Research Centre \\
Tampere University \\
Tampere, Finland \\
\texttt{kestutis.baltakys@tuni.fi}
\And
Hao Dong \\
Data Science Research Centre \\
Tampere University \\
Tampere, Finland \\
\texttt{hao.dong@tuni.fi}
\And
Juho Kanniainen \\
Data Science Research Centre \\
Tampere University \\
Tampere, Finland \\
\texttt{juho.kanniainen@tuni.fi}
}

\iclrfinalcopy 
\begin{document}

\maketitle

\begin{abstract}
Forecasting short-horizon mid-price movements from limit order book (LOB)
data is central to algorithmic trading, yet most deep LOB forecasters are
point predictors: they output a direction or a displacement, but never
indicate which of their forecasts can be trusted. We introduce
\emph{UQ-LOB}, a lightweight,
encoder-agnostic uncertainty quantification module that attaches to any pretrained LOB encoder and, in
the spirit of attentive neural processes, conditions each forecast on a
context set of recently completed windows whose outcomes are already
realised. The \emph{UQ-regression} variant outputs a calibrated Gaussian over
the future tick displacement, while the \emph{UQ-classification}
variant outputs a categorical distribution over down/up/stationary. Both
expose a scalar confidence (predicted signal-to-noise ratio or class
probability) that supports selective prediction. On 5.2 billion LOB
events across seven cryptocurrency assets and horizons of 5, 10 and 15
seconds, UQ-regression attains near-nominal 68\% interval coverage, and restricting to the most confident 10\% of predictions
raises directional macro F1 by 0.11--0.15 for UQ-regression and 0.05--0.11
for UQ-classification, at every horizon. On large, economically meaningful
moves, the tightest confidence tier reaches a directional F1 of 0.88 (down)
and 0.83 (up) at the 5-second horizon.
\end{abstract}

\section{Introduction}
\label{sec:intro}

The limit order book (LOB) is the mechanism through which price discovery
occurs in most electronic markets~\citep{gould2013limit}. It aggregates
resting buy (bid) and sell (ask) orders by price level; the best bid and
best ask define the spread, and their midpoint, the mid-price, is the
standard instantaneous valuation of the
asset~\citep{glosten1985bid,kolm2023deep}. An incoming order that crosses
the spread executes immediately against resting liquidity, whereas one that
does not rests in the book at its own price level~\citep{briola2024deep}.
Because thousands of such events arrive every second, the LOB carries
microstructural signal about the next few seconds of price formation that
predictive models seek to
exploit~\citep{sirignano2019universal,zhang2019deeplob,lucchese2024short}.

Deep LOB forecasters are
overwhelmingly \emph{point predictors}: they emit a direction or a
displacement, but not how much that output should be trusted. For a trading
decision this omission is costly. Predictability varies sharply across
market regimes and even across consecutive windows, and every action incurs
fees and slippage, so the practically relevant question is rarely
``which way will the price move?'' but ``is \emph{this} forecast reliable
enough to act on?'', the setting of selective
prediction~\citep{chow1970optimum,elyaniv2010foundations,geifman2017selective}.
Answering it requires input-dependent (aleatoric) uncertainty that is
\emph{calibrated}: an interval claimed to cover 68\% of outcomes should do
so empirically.

Figure~\ref{fig:intro} illustrates the output we target. Instead of a single
value, the model returns a distribution $\mathcal{N}(\mu,\sigma^{2})$ over
the mid-price displacement at $t+h$, and the ratio $|\mu|/\sigma$ provides
a scalar confidence that can be thresholded. We obtain this with 
\emph{UQ-LOB}, a module attached to the hidden representation of a
pretrained LOB encoder. Inspired by conditional and attentive neural
processes~\citep{garnelo2018conditional,kim2019attentive}, UQ-LOB
conditions each prediction on a \emph{context set} of recently completed
windows whose realised outcomes are already known at time $t$, so that the
forecast can adapt to the prevailing regime without any gradient update.

\begin{minipage}[t]{0.65\linewidth}
\vspace{0pt}
\centering
\includegraphics[width=\linewidth]{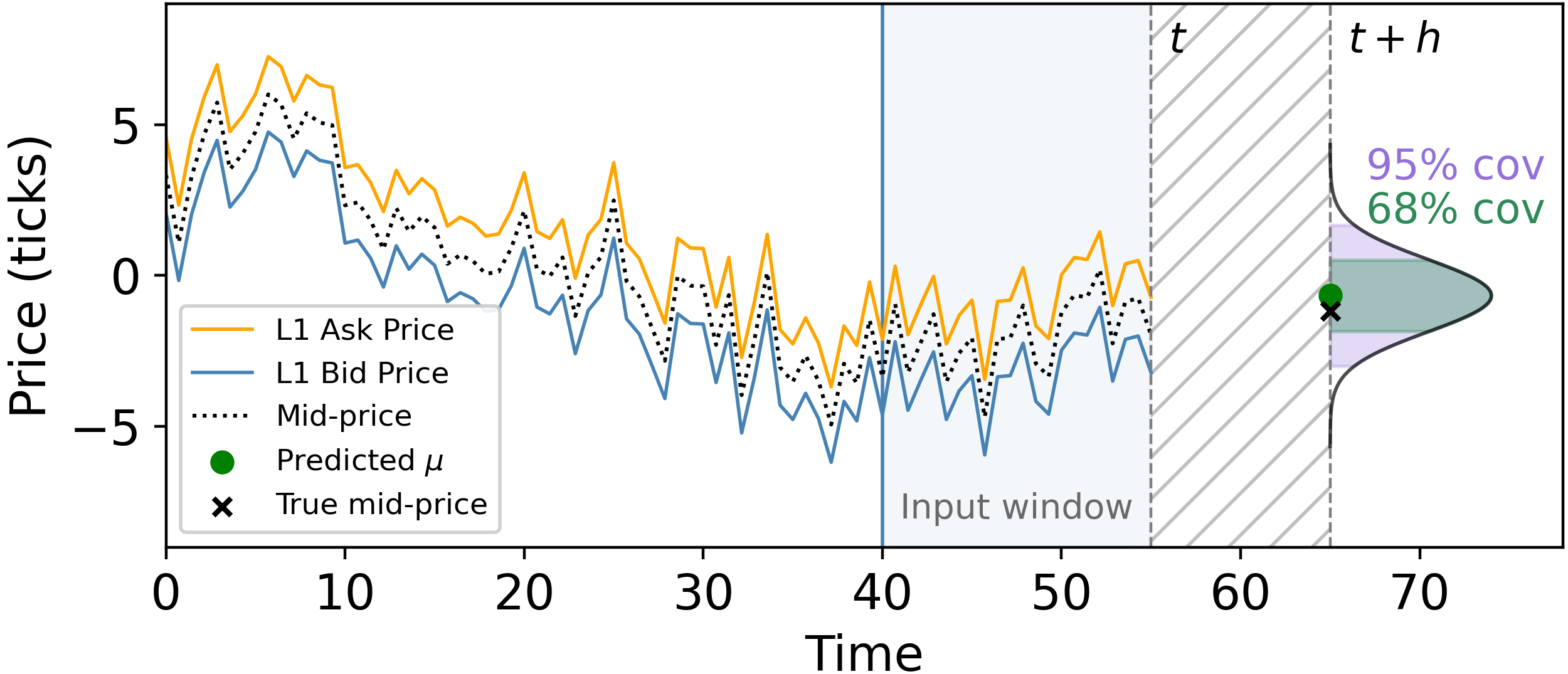}
\vspace{-0.7cm}
\captionof{figure}{Confidence-aware mid-price forecasting. The model sees
only the input window ending at $t$ (shaded) and returns a Gaussian over
the displacement at $t+h$; its 68\% and 95\% intervals give a thresholdable
measure of how much the forecast should be trusted.}
\label{fig:intro}
\end{minipage}%
\hfill
\begin{minipage}[t]{0.33\linewidth}
\vspace{0pt}
To the best of our knowledge, no prior work equips mid-price forecasting
with calibrated, input-dependent uncertainty of this kind. We instantiate
UQ-LOB as a \emph{UQ-regression} variant, which outputs a Gaussian over
the tick displacement, and a \emph{UQ-classification} variant, which
outputs a categorical distribution over the three directional classes
(down, up, stationary), and we compare the two under matched
confidence gating. Our contributions are:
\end{minipage}

\begin{itemize}
    \item \textbf{An encoder-agnostic, in-context UQ head.} An attentive
    head that conditions on the realised outcomes of recent windows and
    produces a calibrated Gaussian $\mathcal{N}(\mu,\sigma^{2})$ over tick
    displacement, with confidence given by its signal-to-noise ratio
    $|\mu|/\sigma$. Its objective combines a heteroscedastic likelihood
    with magnitude weighting, a directional margin, and a differentiable
    magnitude-conditional calibration penalty that turns the post-hoc
    calibration metrics of \citet{levi2022evaluating} into a training
    signal.
    \item \textbf{A classification counterpart on the same trunk}, trained on
    the three-class label with confidence given by its maximum class
    probability, enabling a controlled comparison between continuous and
    discrete targets at identical parameter count.
    \item \textbf{An evaluation protocol for selective LOB prediction} based
    on \emph{directional macro F1}, which removes the trivial effect of the
    stationary class under gating, a magnitude-restricted analysis on
    large moves, and prior-matched random baselines. On 5.2 billion events
    across seven assets and three horizons, confidence gating improves
    directional reliability at every horizon, and the regression head
    matches the dedicated classifier at the top confidence decile while
    additionally providing a magnitude and a calibrated interval.
\end{itemize}

\section{Related Work}
\label{sec:related_work}

\textbf{LOB forecasting.} A large share of deep learning work on LOB data
targets next-message or next-event prediction~\citep{nagy2023generative},
an autoregressive formulation in which errors compound as the effective
horizon grows. The FI-2010 benchmark~\citep{ntakaris2018benchmark}
standardised a second line of work that predicts the direction of the
mid-price over a horizon of $k$ future events as a three-class
label~\citep{zhang2019deeplob,xiao2025lit}. Architectures for this task
have progressed from CNNs~\citep{tsantekidis2017forecasting} and
LSTMs~\citep{tsantekidis2020using} to bilinear and attention-based
models~\citep{tran2018temporal,wallbridge2020transformers}, pretrained
Transformers~\citep{xiao2025lit}, and BERT-style encoders trained on
Level-2/Level-3 event streams~\citep{linna2025lobert}. All of these,
including work that treats the displacement as a regression
target~\citep{yang5269651type}, are point predictors with no accompanying
measure of predictive confidence.

\textbf{Uncertainty in LOB forecasting.} Quantile
regression~\citep{zhang2019extending} jointly models several return
quantiles and avoids quantile crossing, but targets ask- and bid-side
quantiles separately and fixes the quantile levels in advance rather than
yielding a single predictive distribution. BDLOB~\citep{zhang2019bdlob}
applies dropout variational inference~\citep{gal2016dropout} to a LOB CNN
and shows that the resulting posterior predictive uncertainty can be used
for position sizing and for avoiding unnecessary trades;
\citet{magris2023bayesian} obtain predictive distributions over class
probabilities from a Bayesian bilinear network. Both derive uncertainty
from a distribution over network \emph{weights} (epistemic uncertainty)
and require multiple stochastic passes at inference, whereas our head
predicts input-dependent (aleatoric) uncertainty in a single forward pass
and conditions it on recently realised outcomes.

\textbf{Calibration, selective prediction, and neural processes.}
Heteroscedastic Gaussian regression~\citep{nix1994estimating,kendall2017uncertainties}
provides aleatoric uncertainty but is prone to \emph{loss
attenuation}~\citep{seitzer2022pitfalls,stirn2023faithful}; post-hoc
recalibration~\citep{guo2017calibration,kuleshov2018accurate}, deep
ensembles~\citep{lakshminarayanan2017simple}, and conformal
methods~\citep{angelopoulos2023conformal} improve calibration but do not
by themselves make it input-dependent, and predictive uncertainty is known
to degrade under distribution shift~\citep{ovadia2019can}, which is
endemic to financial data. Selective prediction with a reject
option~\citep{chow1970optimum,elyaniv2010foundations,geifman2017selective}
and the maximum softmax probability as a confidence
score~\citep{hendrycks2016baseline} form the evaluation lens we adopt.
Finally, neural processes~\citep{garnelo2018neural} and their conditional
and attentive variants~\citep{garnelo2018conditional,kim2019attentive}
learn predictive distributions conditioned on an observed context set; we
adapt this in-context formulation to the LOB setting, where the context is
the set of most recently completed windows whose labels are already
realised.

\section{UQ-LOB: The Uncertainty Quantification Head}
\label{sec:method}

\begin{figure*}[t]
  \centering
  \includegraphics[width=0.99\textwidth]{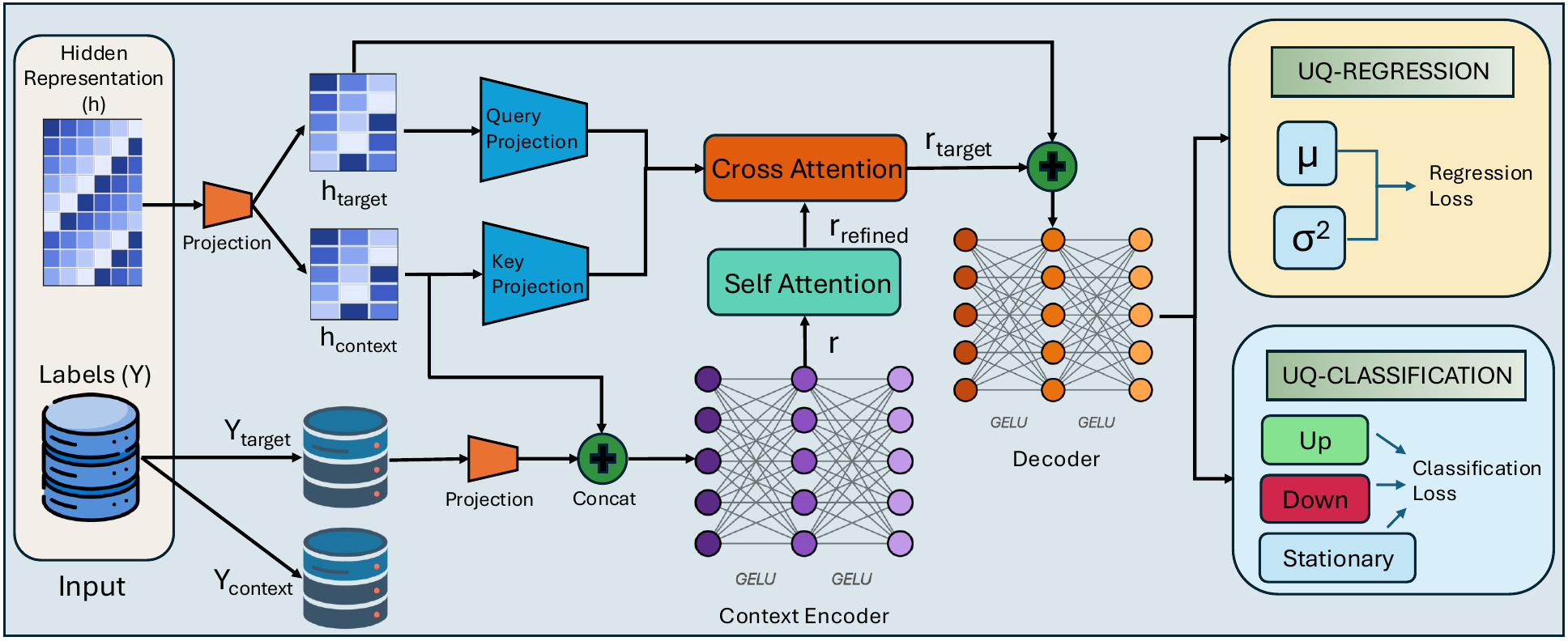}
  \caption{The UQ-LOB. Context and target representations from a
  pretrained encoder are projected into a shared space; each context
  representation is fused with its (scaled) realised label, refined by
  self-attention, and read out by the target through cross-attention. The
  decoder outputs either a Gaussian $(\mu,\sigma^{2})$ over tick
  displacement (UQ-regression) or three class logits (UQ-classification).}
  \label{fig:cnp_arch}
\end{figure*}

Point predictions of mid-price displacement carry no indication of their
own reliability, which is particularly problematic in LOB markets, where
volatility, liquidity and directional momentum shift rapidly across
regimes. UQ-LOB addresses this by producing, for a target window with
hidden representation $\mathbf{h}^{t}$, a full predictive distribution
conditioned on a context set $\mathcal{C}$ of (representation, realised
label) pairs from recently completed windows (Figure~\ref{fig:cnp_arch}).
The head operates on the encoder's hidden representation rather than on raw
market features, so any pretrained LOB encoder can serve as its starting
point (Section~\ref{sec:training_setup}).

\subsection{Problem Setup and Labels}
\label{sec:labels}

The event stream of each asset is segmented into non-overlapping windows of
$L=512$ events (Section~\ref{subsec:input_representation}). Let $t$ denote
the timestamp of the last event of a window, i.e.\ the model's present
moment. To reflect execution latency, the start price $p_{\mathrm{start}}$
is the mid-price at the first event after a small fixed delay $d$ past $t$,
and the end price $p_{\mathrm{end}}$ is the mean mid-price over the
$k=10$ events immediately preceding $t_{\mathrm{start}}+h$, which damps
single-tick noise without washing out genuine displacement at the tick
rates of our data. The regression target is the raw tick displacement
$y_h(t)=p_{\mathrm{end}}-p_{\mathrm{start}}$. The classification label
thresholds the same displacement at $\pm\delta_h(t)$, a volatility-scaled
cutoff derived from the training-split log-return distribution of each
asset and horizon and converted to ticks at the current price level
(Appendix~\ref{app:labelling}). All labels are computed independently per
asset and horizon.

\subsection{In-Context Conditioning}
\label{sec:context}

The head follows conditional neural processes
(CNPs)~\citep{garnelo2018conditional}, which cast regression as an
in-context learning problem: rather than learning a fixed input--output
map, a CNP summarises an observed context set
$\mathcal{C}=\{(\mathbf{x}_i,y_i)\}_{i=1}^{C}$ into a representation
$\mathbf{r}(\mathcal{C})$ and predicts
$p(y^{*}\mid\mathbf{x}^{*},\mathcal{C})=p(y^{*}\mid\mathbf{x}^{*},\mathbf{r}(\mathcal{C}))$.
The original CNP aggregates context points by a mean, which weighs every
point equally; attentive neural processes~\citep{kim2019attentive} let the
query attend selectively to the most relevant points, and we adopt this
form.

In our setting the inputs are windows. Let
$\mathbf{X}_i\in\mathbb{R}^{L\times D}$ be a raw window of $L$ events with
$D$ features, encoded by a pretrained encoder into
$\mathbf{h}_i=f_\theta(\mathbf{X}_i)\in\mathbb{R}^{d_h}$. A training
instance consists of $N=C+1$ windows: $C$ context pairs
$\{(\mathbf{h}_i^{c},y_i^{c})\}_{i=1}^{C}$ and a single withheld target
$(\mathbf{h}^{t},y^{t})$. The context set is drawn from the windows that
most recently precede the target, subject to a \emph{causal constraint}:
if a context window ends at $t_i^{c}$, its label is realised at
$e_i^{c}=t_i^{c}+d+h$, and we require
\begin{equation}
    e_i^{c}\;\le\;t \qquad \forall\, i\in\{1,\dots,C\},
    \label{eq:causal}
\end{equation}
so that every context label is known at the moment the target prediction
is made. This is strictly stronger than merely requiring the context and
target input windows not to overlap: it also excludes context windows whose
label horizon runs past $t$, which would leak future price information
into the prediction. The same construction is used at training, validation
and test time, so the test-time predictor only ever conditions on
information available at $t$ (Appendix~\ref{app:context_alg}).

\subsection{Head Architecture}
\label{sec:architecture}

Figure~\ref{fig:cnp_arch} shows the head. Context and target
representations are projected into a shared space by a common linear map
$W_p$: $\tilde{\mathbf{h}}_i^{c}=W_p\mathbf{h}_i^{c}$ and
$\tilde{\mathbf{h}}^{t}=W_p\mathbf{h}^{t}$. Context labels are divided by a
fixed, asset- and horizon-specific scale $\yref$
(Appendix~\ref{app:per-horizon-thresholds}) and projected through a
$\tanh$-bounded linear layer,
$\mathbf{y}_i^{\mathrm{proj}}=\tanh\!\big(W_y\,y_i^{c}/\yref\big)$. Each
projected representation is concatenated with its projected label and
passed through the \emph{context encoder}, a three-layer GELU MLP,
\begin{equation}
    \mathbf{r}_i = \mathrm{MLP}_{\mathrm{enc}}\big(\left[\tilde{\mathbf{h}}_i^{c} \,;\, \mathbf{y}_i^{\mathrm{proj}}\right]\big), \qquad i = 1,\dots,C.
    \label{eq:ctx_enc}
\end{equation}
Self-attention over $\{\mathbf{r}_i\}_{i=1}^{C}$ then lets each context
point re-weigh itself by inter-context relevance, yielding a refined set
$\{\mathbf{r}_i^{\mathrm{ref}}\}_{i=1}^{C}$. The target queries this
refined context by cross-attention: the projected target embedding is
mapped to a \emph{query} and the projected context embeddings to
\emph{keys} through learned maps $W_q,W_k$, while the refined
representations serve directly as \emph{values},
$\mathbf{q}=W_q\tilde{\mathbf{h}}^{t}$,
$\mathbf{k}_i=W_k\tilde{\mathbf{h}}_i^{c}$,
$\mathbf{v}_i=\mathbf{r}_i^{\mathrm{ref}}$, giving
\begin{equation}
    \mathbf{r}^{t} = \mathrm{softmax}\!\left(\frac{\mathbf{q}\,\mathbf{K}^{\top}}{\sqrt{d_r}}\right)\mathbf{V},
    \label{eq:cross_attn}
\end{equation}
where $\mathbf{K},\mathbf{V}$ stack the keys and values row-wise.
Separating keys from values lets the target decide \emph{which} context
windows are relevant from their representation alone, while retrieving the
label-informed summary of those windows: recent, similarly structured LOB
states receive higher weight than dissimilar ones.

The attended summary is concatenated with the target's own projected
embedding and passed to a \emph{decoder}, a five-layer GELU MLP, which thus
sees both the inferred market context and the target's own features. The
UQ-regression decoder outputs
\begin{equation}
    [\mu,\, \nu] = \mathrm{MLP}_{\mathrm{dec}}\big(\left[\mathbf{r}^{t} \,;\, \tilde{\mathbf{h}}^{t}\right]\big), \qquad
    \sigma^{2} = \yref^{2}\cdot \mathrm{softplus}(\nu) + \epsilon,
    \label{eq:reg_dec}
\end{equation}
giving $p(y^{t}\mid\mathbf{h}^{t},\mathcal{C})=\mathcal{N}(\mu,\sigma^{2})$
with $\epsilon$ a small constant for numerical stability. The
UQ-classification decoder outputs three logits
$[\ell_{\mathrm{down}},\ell_{\mathrm{up}},\ell_{\mathrm{stat}}]=
\mathrm{MLP}_{\mathrm{dec}}^{\mathrm{cls}}([\mathbf{r}^{t};\tilde{\mathbf{h}}^{t}])$.
The two variants share the encoder, projection and attention trunk and
differ only in the decoder (Appendix~\ref{app:param_count}).

\subsection{Training Objectives}
\label{sec:objectives}

\textbf{UQ-regression.} The regression head minimises a weighted sum of
four terms, each computed on target windows only:
\begin{equation}
    \mathcal{L}_{\mathrm{reg}} = \lambda_1 \mathcal{L}_{\mathrm{NLL}} + \lambda_2 \mathcal{L}_{\mathrm{WMAE}} + \lambda_3 \mathcal{L}_{\mathrm{dir}} + \lambda_4 \mathcal{L}_{\mathrm{calib}}.
    \label{eq:total_loss_reg}
\end{equation}
\emph{(i) Heteroscedastic NLL}~\citep{nix1994estimating,kendall2017uncertainties},
$\mathcal{L}_{\mathrm{NLL}}=\mathbb{E}\big[\tfrac{1}{2}\log(2\pi\sigma^{2})+(y-\mu)^{2}/(2\sigma^{2})\big]$.
Letting $\sigma$ depend on the input allows the network to temper the
residual of hard examples by inflating $\sigma$ (\emph{loss
attenuation}). \citet{seitzer2022pitfalls} show this can happen before
$\mu$ has converged: since
$\partial\mathcal{L}_{\mathrm{NLL}}/\partial\mu\propto1/\sigma^{2}$, an
inflated $\sigma$ starves the gradient reaching $\mu$ and biases $\mu$
toward small magnitudes.
\emph{(ii) Magnitude-weighted MAE}, which counteracts this bias by
up-weighting large displacements,
\begin{equation}
    \mathcal{L}_{\mathrm{WMAE}} = \mathbb{E}\!\left[ w(y)\,\frac{|\mu - y|}{\yref} \right],
    \qquad
    w(y) = \min\!\left(1 + \left(\frac{|y|}{\yref}\right)^{p},\ w_{\max}\right),
    \label{eq:wmae}
\end{equation}
with the clamp $w_{\max}$ bounding the influence of extreme outliers
($p=4$, $w_{\max}=40$ throughout).
\emph{(iii) Directional loss.} Neither of the above targets the discrete
up/down/stationary boundary $\delta_i$ used at evaluation time. Writing
$\tilde\mu_i=\mu_i/\yref$, we add
\begin{equation}
    \mathcal{L}_{\mathrm{dir}} = \mathbb{E}\!\left[
        \ind\!\left[|y_i|\geq\delta_i\right]\big(\tilde\mu_i - \mathrm{sign}(y_i)\,m\big)^{2}
        \; + \;
        \ind\!\left[|y_i|<\delta_i\right]\big|\tilde\mu_i - \mathrm{sign}(y_i)\,m\big|
    \right],
    \label{eq:dir}
\end{equation}
where $\delta_i$ is the asset-, horizon- and window-specific threshold of
Appendix~\ref{subsubsec:compare_reg_and_cls} and $m=1$ is a fixed margin.
Directional examples receive a quadratic pull toward a confidently signed
target $\pm m$; stationary examples only a linear one, so their (weaker,
bounded-gradient) sign signal is retained without a strong pull in
magnitude. This class-conditional quadratic/linear switching is
structurally related to the reverse-Huber (berHu)
loss~\citep{zwald2012berhu,laina2016deeper}, which switches on residual
magnitude rather than on the true class.
\emph{(iv) Magnitude-conditional calibration loss.} The three terms above
shape $\mu$; none directly corrects $\sigma$. We therefore add
\begin{equation}
    \mathcal{L}_{\mathrm{calib}} = \frac{1}{B}\sum_{b=1}^{B}
    \left( \mathbb{E}_{i\in \mathcal{B}_b}\!\left[\frac{|\mu_i - y_i|}{\sigma_i}\right] - \sqrt{\tfrac{2}{\pi}} \right)^{2},
    \label{eq:calib}
\end{equation}
where $\{\mathcal{B}_b\}_{b=1}^{B}$ partitions the batch into $B$ quantile
bins of $|\mu|$ and $\sqrt{2/\pi}=\mathbb{E}|Z|$, $Z\sim\mathcal{N}(0,1)$,
is the expected standardised residual of a calibrated Gaussian. Unlike
remedies that reweight the NLL globally~\citep{seitzer2022pitfalls} or
decouple the mean and variance branches
architecturally~\citep{stirn2023faithful}, $\mathcal{L}_{\mathrm{calib}}$
turns the post-hoc calibration metrics of \citet{levi2022evaluating} into a
differentiable penalty, binning on predicted magnitude so that calibration
is enforced separately for small and large predicted moves. All loss
hyperparameters are listed in Appendix~\ref{app:hyperparams}.

\textbf{UQ-classification.} The classification head minimises a
class-weighted cross-entropy over the three-class label,
$\mathcal{L}_{\mathrm{cls}}=-\sum_{c=0}^{2}w_c\,\ind[y_{\mathrm{cls}}=c]\log p_c$,
with $w_c\propto1/f_c$ the inverse training-split frequency of class $c$,
normalised so that $\sum_c w_c=3$, which counteracts the class imbalance
documented in Appendix~\ref{app:data_characteristics}.

\subsection{Confidence Scores and Selective-Prediction Protocol}
\label{sec:protocol}

\textbf{Confidence.} For UQ-regression we use the predicted
signal-to-noise ratio $\mathrm{SNR}=|\mu|/\sigma$: a large predicted
displacement paired with a small predicted uncertainty is both worth
acting on and sharply estimated. For UQ-classification we use the softmax
confidence $\max_c p_c$~\citep{hendrycks2016baseline}. Gating at
percentile $q$ retains the $(100-q)\%$ most confident predictions of each
asset--horizon test set and evaluates only on those; $q=0$ is the full
test set.

\textbf{Directional macro F1.} To compare the heads on a common footing,
the regression mean $\mu$ is mapped to a three-class label with the same
rule as the ground truth, using a threshold multiplier calibrated on a
held-out split so that the predicted stationary proportion matches the true
one (Appendix~\ref{subsubsec:compare_reg_and_cls}). Under gating, the
stationary class behaves degenerately: confident predictions are large in
magnitude and therefore predominantly directional, so stationary F1
collapses while down/up F1 rise (Table~\ref{tab:per-class-f1-snr-5s}),
and three-class macro F1 conflates the two effects. We therefore report
\emph{directional macro F1}, the mean of the down and up F1 scores, and
provide a prior-matched reference: a classifier predicting at random with
the true class priors attains an expected F1 of $p_c$ for class $c$, so its
directional macro F1 is $(1-p_{\mathrm{stat}})/2$. Because gating changes
the class composition of the retained subset, this reference is only
valid for the full test set and rises under gating; the large-move analysis
of Section~\ref{sec:large_moves} fixes it at 0.5 by construction.

\textbf{Calibration and magnitude-weighted fit.} We report the empirical
coverage of the nominal 68\% and 95\% intervals,
$\mathrm{cov}_{68}=\Pr[|y-\mu|\le\sigma]$ and
$\mathrm{cov}_{95}=\Pr[|y-\mu|\le1.96\sigma]$, and the negative log
predictive density (NLPD). Point fit is summarised by a magnitude-weighted
$R^{2}$,
\begin{equation}
    R^2_w = 1 - \frac{\sum_i w_i (y_i - \mu_i)^2}{\sum_i w_i (y_i - \bar{y}_w)^2},
    \qquad
    \bar{y}_w = \frac{\sum_i w_i y_i}{\sum_i w_i},
    \label{eq:weighted_r2}
\end{equation}
with $w_i=w(y_i)$ from Equation~\ref{eq:wmae}. Large moves are those that
cross the stationary threshold and, after fees, are the only ones a trader
can profit from; \WR{} therefore measures fit on the moves that matter
rather than on the small fluctuations that dominate standard $R^{2}$.

\section{Experimental Setup}
\label{sec:setup}

\subsection{Dataset}
\label{sec:dataset}

We use Level-2 and Level-3 order book data collected from the Kraken
cryptocurrency exchange through its public market data API for seven
USD-quoted assets: Bitcoin (BTC), Ethereum (ETH), Solana (SOL), Litecoin
(LTC), Dogecoin (DOGE), Sui (SUI) and Bittensor (TAO). The data span
November 2025 to February 2026 and contain approximately 5.2 billion
events over 600 asset-days, with event rates from 39 to 156 events/s and
mean inter-event times from 6.4 to 26.0\,ms
(Table~\ref{tab:dataset-stats}, Appendix~\ref{app:data_characteristics}).
For every asset, data up to 18 January 2026 are used for training,
19--31 January for validation and 1--15 February for testing (about 68\%,
15\% and 17\% of asset-days). The validation period is split into two
equal halves for model selection and for calibrating the threshold
multiplier of Appendix~\ref{subsubsec:compare_reg_and_cls}; the test
period is strictly out-of-sample and chronologically after both. Because
no data-sharing agreement is in place with the exchange, the raw data
cannot be redistributed; we release code, all label thresholds and scale
values, and the dataset statistics needed to reconstruct the pipeline on
comparable data (Reproducibility Statement).

\subsection{Input Representation}
\label{subsec:input_representation}

Following \citet{linna2025lobert}, each event is encoded as a discrete
token from a vocabulary of $|\mathcal{V}|=439$ entries built from five
categorical attributes (event type, side, discretised volume, discretised
price distance from the opposing best quote, and a simultaneity flag);
events deeper than $L_{\max}=10$ levels are discarded. Each event also
carries seven continuous features. Three follow
\citet{linna2025lobert}: log inter-event time, log tick distance from the
opposing best quote and log relative volume. Four capture directional
order-flow pressure at the top of the book. Building on the order flow
imbalance of \citet{cont2014price}, the depth-normalised flow imbalance
\begin{equation}
    \mathrm{DNFI}_t =
    \frac{\Delta Q^b_{1,t} - \Delta Q^a_{1,t}}{Q^b_{1,t} + Q^a_{1,t}}
    \in [-1,+1],
    \qquad
    \mathrm{QI}_t =
    \frac{Q^b_{1,t} - Q^a_{1,t}}{Q^b_{1,t} + Q^a_{1,t}}
    \in [-1,+1],
    \label{eq:dnfi}
\end{equation}
tracks the net change in best-level queue sizes $Q^{b}_{1,t},Q^{a}_{1,t}$
relative to resting depth ($\Delta$ denotes the change from the previous
event); two cumulative sums of DNFI over the preceding 50 and 200 events
capture short- and medium-term flow momentum; and the queue imbalance
$\mathrm{QI}_t$ complements the flow signal with a static snapshot of
resting-liquidity asymmetry. The stream is segmented into non-overlapping
windows of $L=512$ events, each forming one model input
(Appendix~\ref{app:input_rep}).

\subsection{Encoder, Training and Model Size}
\label{sec:training_setup}

As encoder we use D-TABL, a deeper variant of the temporal
attention-augmented bilinear network of \citet{tran2018temporal} adapted to
our $512$-event windows (Appendix~\ref{app:DTABL}). It is first pretrained
on the three-class label alone, independently for each of the 21
asset--horizon pairs. UQ-LOB is then attached to the pretrained
encoder and the whole model is fine-tuned end-to-end with
Equation~\ref{eq:total_loss_reg} (UQ-regression) or the weighted
cross-entropy (UQ-classification). Each training instance consists of
$N=16$ windows, $C=15$ context windows and one withheld target, and each
step processes a batch of 16 such instances. The encoder has 0.27M
parameters; the shared projection and context/attention trunk adds 0.79M,
and each decoder 0.17M, so both variants total 1.23M parameters
(Table~\ref{tab:param-count}). Training all 42 models requires about 210
GPU-hours on a single NVIDIA GH200; full details and hyperparameters are
given in Appendices~\ref{app:training_strategy} and~\ref{app:hyperparams}.

\section{Results}
\label{sec:results}

\subsection{Calibration of the Regression Head}
\label{sec:calibration}

\noindent
\begin{minipage}[t]{0.55\linewidth}
\vspace{0pt}
\centering
\includegraphics[width=\linewidth]{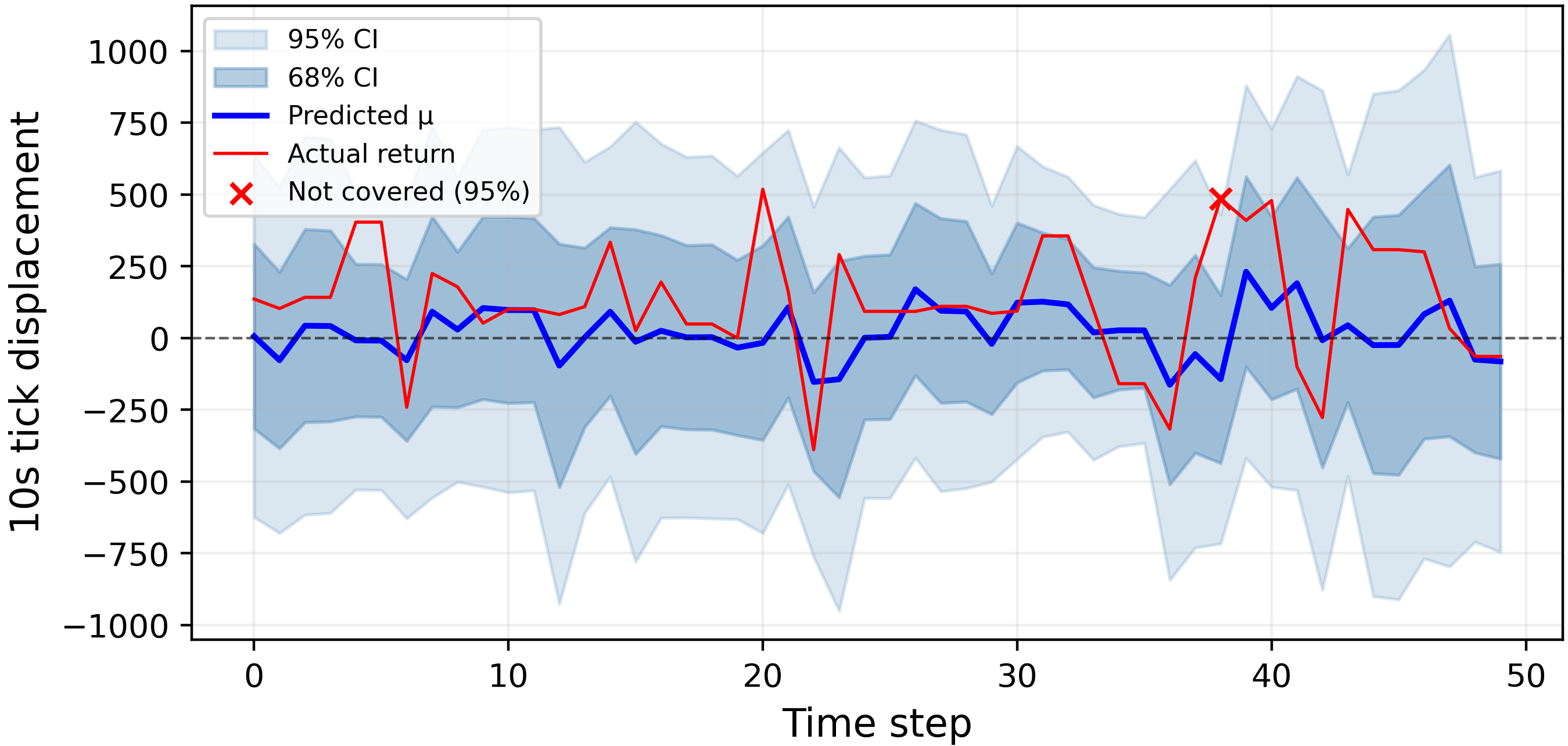}
\vspace{-0.7cm}
\captionof{figure}{UQ-regression on consecutive BTC-USD test windows at the
10\,s horizon: predicted $\mu$ with 68\% and 95\% intervals against the
realised displacement; crosses mark realisations outside the 95\%
interval.}
\label{fig:calibration}
\end{minipage}%
\hfill
\begin{minipage}[t]{0.42\linewidth}
\vspace{0pt}
Figure~\ref{fig:calibration} gives a qualitative view of the predictive
distribution, and Table~\ref{tab:uncertainty-calibration} quantifies it
across horizons. Empirical 68\% coverage is 68.1--68.2\%, essentially
nominal, and it is stable across horizons. The 95\% interval, however,
covers only 90.4--90.7\% of outcomes: the residual distribution is heavier
tailed than a Gaussian, and Equation~\ref{eq:calib} constrains only the
first absolute moment of the standardised residual, so it enforces the
one-sigma level but not the tails. We return to this in
Section~\ref{sec:limitations}.
\end{minipage}

\subsection{Confidence Gating}
\label{sec:gating}

\begin{figure*}[t]
  \centering
  \includegraphics[width=0.9\textwidth]{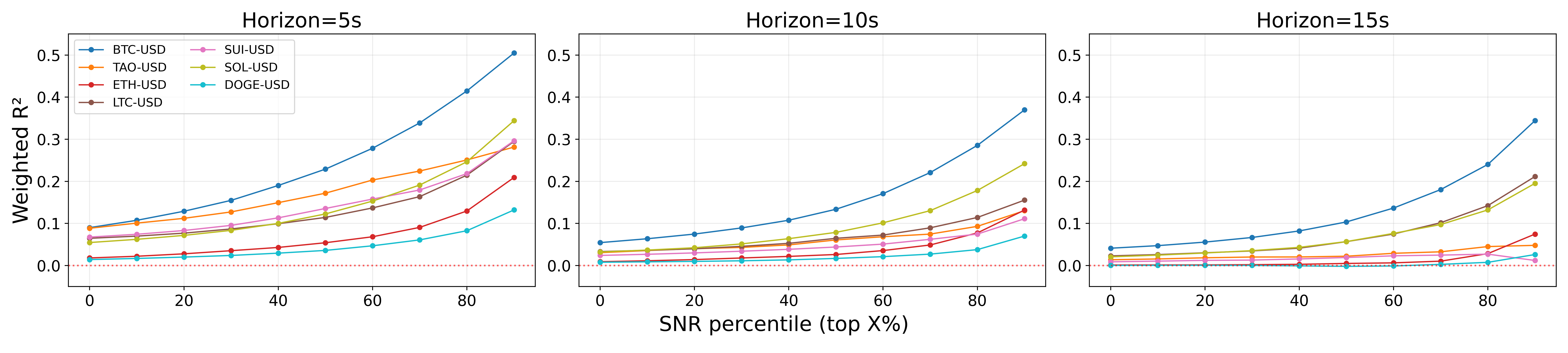}
    \caption{Magnitude-weighted $R^{2}$ (Equation~\ref{eq:weighted_r2}) of
    the UQ-regression head as a function of the SNR percentile retained,
    per asset and horizon.}
  \label{fig:regression-classification-results}
\end{figure*}

\begin{table*}[t]
\caption{Directional macro F1 (mean of down and up F1; stationary excluded)
as a function of the confidence percentile retained, for UQ-regression
(SNR-gated) and UQ-classification (softmax-gated), pooled over the seven
assets. Row $0$ is the full test set. A prior-matched random classifier
attains $(1-p_{\mathrm{stat}})/2$ on the full test set: 0.337 at 5\,s
($p_{\mathrm{stat}}=0.326$), 0.350 at 10\,s ($p_{\mathrm{stat}}=0.300$) and
0.362 at 15\,s ($p_{\mathrm{stat}}=0.276$); because gating changes the class
mix of the retained subset, this full-set reference does not apply to the
gated rows (Section~\ref{sec:gating}).}
\label{tab:directional-f1-table}
\resizebox{\textwidth}{!}{%
\begin{tabular}{l | rrr | rrr | rrr | rrr | rrr | rrr}
\toprule
\textbf{H} & \multicolumn{6}{c|}{\textbf{5s}} & \multicolumn{6}{c|}{\textbf{10s}} & \multicolumn{6}{c}{\textbf{15s}} \\
\textbf{} & \multicolumn{3}{c|}{\textnormal{UQ-Regression}} & \multicolumn{3}{c|}{\textnormal{UQ-Classification}} & \multicolumn{3}{c|}{\textnormal{UQ-Regression}} & \multicolumn{3}{c|}{\textnormal{UQ-Classification}} & \multicolumn{3}{c|}{\textnormal{UQ-Regression}} & \multicolumn{3}{c}{\textnormal{UQ-Classification}} \\
\textbf{Conf.} & \textbf{F1(\dn)} & \textbf{F1(\up)} & Mean
& \textbf{F1(\dn)} &  \textbf{F1(\up)} & Mean
& \textbf{F1(\dn)} &  \textbf{F1(\up)} & Mean
& \textbf{F1(\dn)} &  \textbf{F1(\up)} & Mean
& \textbf{F1(\dn)} &  \textbf{F1(\up)} & Mean
& \textbf{F1(\dn)} &  \textbf{F1(\up)} & Mean \\
\midrule
0  & \cellcolor{red!10}0.45 & \cellcolor{red!35}0.39 & \cellcolor{red!20}0.42 & \cellcolor{green!15}0.48 & \cellcolor{red!18}0.42 & \cellcolor{red!10}0.45 & \cellcolor{red!12}0.43 & \cellcolor{red!38}0.39 & \cellcolor{red!25}0.41 & \cellcolor{green!5}0.46 & \cellcolor{red!10}0.43 & \cellcolor{red!10}0.45 & \cellcolor{red!5}0.44 & \cellcolor{red!40}0.36 & \cellcolor{red!30}0.40 & \cellcolor{green!12}0.47 & \cellcolor{red!18}0.42 & \cellcolor{red!5}0.45 \\
10 & \cellcolor{green!10}0.47 & \cellcolor{red!25}0.41 & \cellcolor{red!5}0.44 & \cellcolor{green!25}0.49 & \cellcolor{red!15}0.43 & \cellcolor{green!5}0.46 & \cellcolor{green!1}0.45 & \cellcolor{red!25}0.41 & \cellcolor{red!10}0.43 & \cellcolor{green!10}0.47 & \cellcolor{red!5}0.44 & \cellcolor{green!2}0.45 & \cellcolor{green!8}0.46 & \cellcolor{red!25}0.38 & \cellcolor{red!15}0.42 & \cellcolor{green!20}0.48 & \cellcolor{red!15}0.43 & \cellcolor{green!4}0.46 \\
20 & \cellcolor{green!22}0.49 & \cellcolor{red!10}0.43 & \cellcolor{green!8}0.46 & \cellcolor{green!32}0.50 & \cellcolor{red!12}0.43 & \cellcolor{green!10}0.47 & \cellcolor{green!14}0.47 & \cellcolor{red!12}0.43 & \cellcolor{green!2}0.45 & \cellcolor{green!17}0.48 & \cellcolor{red!2}0.44 & \cellcolor{green!6}0.46 & \cellcolor{green!20}0.48 & \cellcolor{red!10}0.40 & \cellcolor{red!5}0.44 & \cellcolor{green!25}0.49 & \cellcolor{red!12}0.43 & \cellcolor{green!8}0.46 \\
30 & \cellcolor{green!32}0.51 & \cellcolor{red!2}0.45 & \cellcolor{green!18}0.48 & \cellcolor{green!38}0.51 & \cellcolor{red!10}0.43 & \cellcolor{green!14}0.47 & \cellcolor{green!22}0.49 & \cellcolor{red!2}0.45 & \cellcolor{green!10}0.47 & \cellcolor{green!22}0.48 & \cellcolor{green!1}0.45 & \cellcolor{green!10}0.47 & \cellcolor{green!30}0.50 & \cellcolor{red!05}0.41 & \cellcolor{green!3}0.46 & \cellcolor{green!30}0.50 & \cellcolor{red!10}0.43 & \cellcolor{green!11}0.47 \\
40 & \cellcolor{green!38}0.52 & \cellcolor{green!8}0.47 & \cellcolor{green!25}0.49 & \cellcolor{green!45}0.52 & \cellcolor{red!8}0.44 & \cellcolor{green!18}0.48 & \cellcolor{green!28}0.50 & \cellcolor{green!5}0.46 & \cellcolor{green!18}0.48 & \cellcolor{green!28}0.49 & \cellcolor{green!3}0.45 & \cellcolor{green!14}0.47 & \cellcolor{green!35}0.51 & \cellcolor{red!02}0.42 & \cellcolor{green!10}0.47 & \cellcolor{green!35}0.51 & \cellcolor{red!8}0.44 & \cellcolor{green!14}0.47 \\
50 & \cellcolor{green!45}0.53 & \cellcolor{green!15}0.48 & \cellcolor{green!30}0.50 & \cellcolor{green!50}0.53 & \cellcolor{red!2}0.45 & \cellcolor{green!25}0.49 & \cellcolor{green!33}0.51 & \cellcolor{green!10}0.47 & \cellcolor{green!22}0.49 & \cellcolor{green!35}0.50 & \cellcolor{green!5}0.45 & \cellcolor{green!18}0.48 & \cellcolor{green!40}0.52 & \cellcolor{green!01}0.42 & \cellcolor{green!13}0.47 & \cellcolor{green!40}0.52 & \cellcolor{red!7}0.44 & \cellcolor{green!18}0.48 \\
60 & \cellcolor{green!50}0.54 & \cellcolor{green!22}0.49 & \cellcolor{green!35}0.51 & \cellcolor{green!58}0.55 & \cellcolor{green!3}0.46 & \cellcolor{green!30}0.50 & \cellcolor{green!38}0.52 & \cellcolor{green!15}0.48 & \cellcolor{green!28}0.50 & \cellcolor{green!40}0.51 & \cellcolor{green!10}0.46 & \cellcolor{green!22}0.48 & \cellcolor{green!45}0.53 & \cellcolor{green!05}0.43 & \cellcolor{green!16}0.48 & \cellcolor{green!48}0.53 & \cellcolor{red!5}0.44 & \cellcolor{green!22}0.48 \\
70 & \cellcolor{green!58}0.55 & \cellcolor{green!30}0.50 & \cellcolor{green!42}0.52 & \cellcolor{green!65}0.56 & \cellcolor{green!10}0.47 & \cellcolor{green!38}0.51 & \cellcolor{green!45}0.53 & \cellcolor{green!22}0.49 & \cellcolor{green!34}0.51 & \cellcolor{green!48}0.52 & \cellcolor{green!18}0.47 & \cellcolor{green!28}0.49 & \cellcolor{green!52}0.54 & \cellcolor{green!15}0.43 & \cellcolor{green!20}0.48 & \cellcolor{green!55}0.54 & \cellcolor{red!2}0.45 & \cellcolor{green!28}0.49 \\
80 & \cellcolor{green!68}0.57 & \cellcolor{green!42}0.52 & \cellcolor{green!52}0.54 & \cellcolor{green!75}0.58 & \cellcolor{green!20}0.48 & \cellcolor{green!48}0.53 & \cellcolor{green!55}0.54 & \cellcolor{green!30}0.50 & \cellcolor{green!42}0.52 & \cellcolor{green!58}0.53 & \cellcolor{green!20}0.47 & \cellcolor{green!34}0.50 & \cellcolor{green!60}0.55 & \cellcolor{green!20}0.44 & \cellcolor{green!26}0.49 & \cellcolor{green!62}0.55 & \cellcolor{green!4}0.45 & \cellcolor{green!34}0.50 \\
90 & \cellcolor{green!80}0.60 & \cellcolor{green!55}0.54 & \cellcolor{green!65}0.57 & \cellcolor{green!85}0.61 & \cellcolor{green!35}0.51 & \cellcolor{green!60}0.56 & \cellcolor{green!65}0.57 & \cellcolor{green!40}0.53 & \cellcolor{green!52}0.55 & \cellcolor{green!65}0.55 & \cellcolor{green!15}0.46 & \cellcolor{green!38}0.50 & \cellcolor{green!72}0.57 & \cellcolor{green!30}0.46 & \cellcolor{green!38}0.51 & \cellcolor{green!72}0.57 & \cellcolor{green!20}0.47 & \cellcolor{green!45}0.52 \\
\bottomrule
\end{tabular}%
}
\end{table*}

\begin{figure*}[tp]
    \centering
    \includegraphics[width=0.9\textwidth]{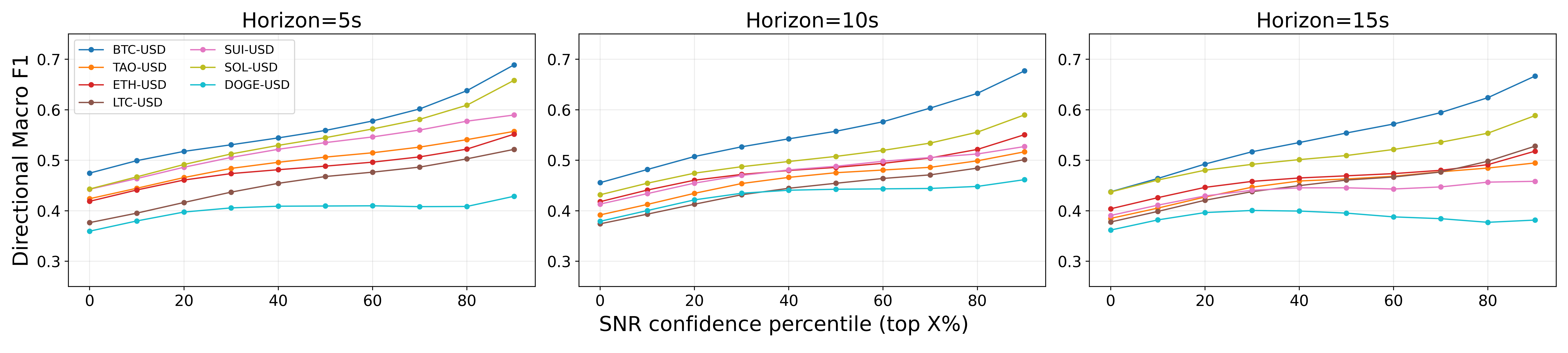}
    \vspace{0.1em}
    \includegraphics[width=0.9\textwidth]{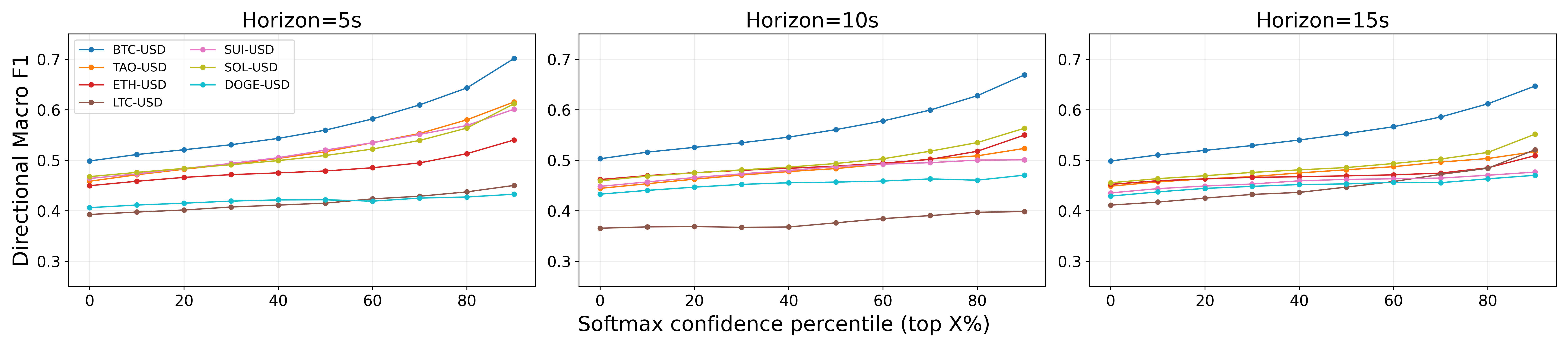}
    \caption{Directional macro F1 as a function of the confidence percentile
    retained, per asset and horizon. Top: UQ-regression gated by SNR.
    Bottom: UQ-classification gated by softmax confidence.}
    \label{fig:confidence-gated-macro-f1-results}
\end{figure*}

\textbf{UQ-regression.} Figure~\ref{fig:regression-classification-results}
plots \WR{} against the SNR percentile retained. For every asset and
horizon, \WR{} rises monotonically as the retained subset is restricted to
higher-SNR predictions, most markedly for BTC-USD (from roughly 9\% on the
full 5\,s test set to over 50\% in the top decile). Three-class accuracy
under SNR gating follows the same pattern
(Figure~\ref{fig:classification-accuracy-results},
Appendix~\ref{app:extended-results}; BTC-USD rises from about 45\% to 57\%
at 5\,s). Table~\ref{tab:directional-f1-table} reports directional macro
F1: between the full test set and the most confident 10\% of predictions
it rises from 0.42 to 0.57 at 5\,s, from 0.41 to 0.55 at 10\,s and from
0.40 to 0.51 at 15\,s, and Figure~\ref{fig:confidence-gated-macro-f1-results}
(top) shows that the increase is monotone over the whole percentile range
for every asset, with DOGE at 15\,s the only mild exception at the highest
tiers. Two caveats apply. Both heads score lower on the up class than on the
down class at every tier; as the asymmetry is shared, it reflects the test
period rather than either head. And since the prior-matched reference
rises with the gate as the retained subset becomes more directional, the
top-decile values should not be read against the full-set reference of
0.34--0.36; the magnitude-restricted analysis below is the cleaner test.
\todo{Add the retained-subset prior-matched reference at each percentile
(one extra row per horizon, or a figure in Appendix C); it is a one-line
computation from the saved predictions.}

\textbf{UQ-classification.} Under softmax gating the same directional
macro F1 rises from 0.45 to 0.56 at 5\,s, 0.45 to 0.50 at 10\,s and 0.45 to
0.52 at 15\,s (Table~\ref{tab:directional-f1-table}), monotonically at
every horizon (Figure~\ref{fig:confidence-gated-macro-f1-results},
bottom). Three-class macro F1 and accuracy show the same behaviour
(Figures~\ref{fig:softmax-vs-f1} and~\ref{fig:classification-accuracy-results},
Appendix~\ref{app:extended-results}): for BTC-USD, macro F1 climbs from
0.47 on the full 5\,s test set to nearly 0.68 in the top decile, and
accuracy from about 48\% to 68\%. Softmax confidence is therefore a
reliable ranking signal for this head as well.

\subsection{Large Moves}
\label{sec:large_moves}

Table~\ref{tab:snr-diracc-large-moves-f1-updown} restricts evaluation to
large true moves, defined per asset and horizon as $|y|>q_{67}$, the upper
tercile of the true displacement magnitude. By construction these examples
are almost never stationary and the down/up split is close to balanced,
so a prior-matched random classifier scores about 0.5 on either class
regardless of the gate, which makes this the cleanest test of whether
confidence identifies correct directional calls. Per-class F1 rises
steadily with SNR at every horizon: at 5\,s, F1(down) increases from 0.69
in the top 50\% to 0.88 in the top 1\%, and F1(up) from 0.63 to 0.83; the
same pattern holds at 10\,s and 15\,s, with a smaller gain on the up class
at 15\,s. This corroborates the gating result of
Table~\ref{tab:directional-f1-table} under an independent, magnitude-based
restriction, and it is exactly the regime, large moves at short horizons,
in which a reliable signal is most valuable after transaction costs.

\noindent
\begin{minipage}[t]{0.62\linewidth}
\vspace{0pt}
\centering
\captionof{table}{Per-class directional F1 of UQ-regression on large true
moves ($|y|>q_{67}$) at increasing SNR tiers.}
\label{tab:snr-diracc-large-moves-f1-updown}
\scriptsize
\setlength{\tabcolsep}{3pt}
\begin{tabular}{l|rr|rr|rr|rr|rr}
\toprule
& \multicolumn{2}{c|}{\textbf{Top 50\%}}
& \multicolumn{2}{c|}{\textbf{Top 30\%}}
& \multicolumn{2}{c|}{\textbf{Top 10\%}}
& \multicolumn{2}{c|}{\textbf{Top 5\%}}
& \multicolumn{2}{c}{\textbf{Top 1\%}} \\
\cmidrule(lr){2-3}
\cmidrule(lr){4-5}
\cmidrule(lr){6-7}
\cmidrule(lr){8-9}
\cmidrule(lr){10-11}
\textbf{H}
& \textbf{F1(\dn)} & \textbf{F1(\up)}
& \textbf{F1(\dn)} & \textbf{F1(\up)}
& \textbf{F1(\dn)} & \textbf{F1(\up)}
& \textbf{F1(\dn)} & \textbf{F1(\up)}
& \textbf{F1(\dn)} & \textbf{F1(\up)} \\
\midrule
5s  & \cellcolor{green!25}0.69 & \cellcolor{green!18}0.63
    & \cellcolor{green!34}0.73 & \cellcolor{green!24}0.67
    & \cellcolor{green!48}0.81 & \cellcolor{green!39}0.75
    & \cellcolor{green!54}0.84 & \cellcolor{green!46}0.79
    & \cellcolor{green!60}0.88 & \cellcolor{green!53}0.83 \\
10s & \cellcolor{green!18}0.65 & \cellcolor{green!12}0.59
    & \cellcolor{green!24}0.68 & \cellcolor{green!18}0.63
    & \cellcolor{green!38}0.75 & \cellcolor{green!30}0.70
    & \cellcolor{green!43}0.78 & \cellcolor{green!35}0.73
    & \cellcolor{green!55}0.85 & \cellcolor{green!46}0.78 \\
15s & \cellcolor{green!18}0.65 & \cellcolor{green!5}0.52
    & \cellcolor{green!24}0.68 & \cellcolor{green!6}0.53
    & \cellcolor{green!37}0.74 & \cellcolor{green!12}0.59
    & \cellcolor{green!41}0.76 & \cellcolor{green!16}0.61
    & \cellcolor{green!48}0.81 & \cellcolor{green!26}0.68 \\
\bottomrule
\end{tabular}
\end{minipage}%
\hfill
\begin{minipage}[t]{0.36\linewidth}
\vspace{0pt}
\centering
\captionof{table}{Calibration of UQ-regression (full test set, pooled over
assets). Calib.\ error: mean over assets of $|\mathrm{cov}_{68}-0.68|$.}
\label{tab:uncertainty-calibration}
\scriptsize
\setlength{\tabcolsep}{4pt}
\begin{tabular}{lrrr}
\toprule
\textbf{Metric} & \textbf{5s} & \textbf{10s} & \textbf{15s} \\
\midrule
Cov.\ 68 (nominal 0.68) & 0.682 & 0.682 & 0.681 \\
Cov.\ 95 (nominal 0.95) & 0.904 & 0.904 & 0.907 \\
Calib.\ error  & 0.048 & 0.048 & 0.045 \\
NLPD           & 5.278 & 5.586 & 5.778 \\
\bottomrule
\end{tabular}
\end{minipage}
\todo{Confirm the definition of ``Calib.\ error'' used in the caption
(per-asset $|\mathrm{cov}_{68}-0.68|$ averaged over assets) matches the
code; if it is ENCE/UCE of Levi et al., say so and cite.}

\subsection{Regression versus Classification Head}
\label{sec:head_comparison}

The two heads share every parameter except the decoder, so
Table~\ref{tab:directional-f1-table} is a controlled comparison of the
two targets. On the full test set the dedicated classifier is stronger by
0.03--0.05 directional macro F1 (0.45 versus 0.40--0.42), as one would
expect from a head trained directly on the evaluation label. Gating closes
this gap: at the top decile the regression head matches or exceeds the
classifier (0.57 versus 0.56 at 5\,s, 0.55 versus 0.50 at 10\,s, 0.51
versus 0.52 at 15\,s), and the gain from gating is larger for SNR than for softmax
confidence at every horizon (+0.11 to +0.15 versus +0.05 to +0.11). Since SNR combines predicted magnitude with
predicted precision, its top tier is enriched for large, well-estimated
moves, exactly the ones on which the regression head is strongest
(Section~\ref{sec:large_moves}). The regression head thus offers, at
identical cost, a calibrated magnitude and interval in addition to a
direction, and loses nothing in directional reliability on the subset a
trader would act on.

\section{Conclusion}
\label{sec:conclusion}

We introduced UQ-LOB, an encoder-agnostic module that turns a
pretrained LOB encoder into a calibrated probabilistic forecaster by
conditioning on the realised outcomes of recently completed windows. Across
seven cryptocurrency assets and three horizons, its regression variant
attains near-nominal 68\% coverage, and gating by its own signal-to-noise
ratio raises directional macro F1 by 0.11--0.15 at the top confidence
decile (0.05--0.11 for the classification variant under softmax gating),
reaching F1 of 0.88 (down) and 0.83 (up) on large 5-second moves. Although
not trained on the discrete label, the regression head matches the
dedicated classifier at the top decile while also supplying a magnitude
and a calibrated interval: a single head can serve both continuous and
directional decisions together with a grounded measure of when to act.

\section{Limitations and Future Work}
\label{sec:limitations}

\textbf{Tail calibration.} The 95\% interval under-covers (90.4--90.7\%),
because the Gaussian likelihood and the first-moment calibration penalty do
not capture the heavy tails of tick displacements; a Student-$t$ likelihood
or post-hoc conformal recalibration of the intervals is a direct extension.
\textbf{Baselines and variance.} Our comparison is between the two UQ heads
on a shared trunk; a comparison against weight-space uncertainty (MC
dropout, deep ensembles) and against gating the base encoder's own
softmax, together with multi-seed error bars and ablations of the four loss
terms and of the context set, would sharpen the attribution of the gains
to in-context conditioning. \textbf{Distribution shift.} The training and
test windows differ markedly in regime for BTC-USD (stationary share 70\%
versus 27\% at 5\,s, Appendix~\ref{subsec:class-proportion-drift}); that
gating still works there is encouraging, but a longer test period spanning
several regimes is needed to establish robustness. \textbf{Economic
evaluation.} We evaluate directional reliability, not profitability; a
trading simulation with realistic execution and fees is required before
any claim about returns. \textbf{Scope.} Models are trained separately per
asset and horizon on a single venue; a single model trained jointly across
horizons, and wider input windows for longer horizons, are natural next
steps.

\subsection*{Ethics Statement}
This work uses only public market data collected through the exchange's
public API; it contains no personal data. The models forecast short-horizon
price movements and could inform trading decisions; we make no claim of
profitability, report all limitations above, and note that acting on such
forecasts carries financial risk. We see no direct negative societal
impact beyond those common to quantitative trading research.

\subsection*{Reproducibility Statement}
We release the complete source code for the D-TABL encoder, both UQ heads,
and the training and evaluation pipelines, including the context
construction of Appendix~\ref{app:context_alg}. The raw order book data
cannot be redistributed (Section~\ref{sec:dataset}); to support
reproduction on comparable Level-2/Level-3 data from any venue we release
all label thresholds and scale values (Table~\ref{tab:per-horizon-thresholds}),
the dataset statistics (Table~\ref{tab:dataset-stats}), and the full
specification of the labelling procedure (Appendix~\ref{app:labelling}),
input representation (Section~\ref{subsec:input_representation} and
Appendix~\ref{app:input_rep}), training strategy
(Appendix~\ref{app:training_strategy}) and hyperparameters
(Appendix~\ref{app:hyperparams}).
\clearpage
\bibliography{iclr2027_conference}
\bibliographystyle{iclr2027_conference}

\clearpage
\appendix

\section{Model Specifications}

\subsection{D-TABL Encoder}
\label{app:DTABL}

\begin{figure}[t]
    \centering
    \includegraphics[width=0.95\linewidth]{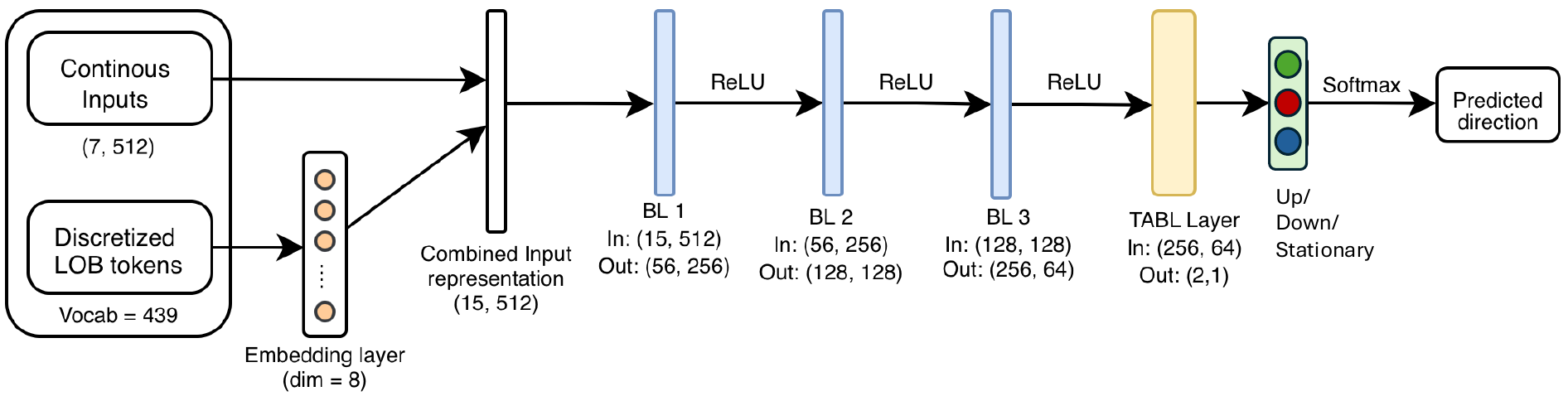}
    \caption{The D-TABL encoder: three bilinear layers followed by a
    temporal-attention-augmented bilinear (TABL) layer; the token embedding
    is concatenated with the continuous features at every time step. The
    UQ head consumes the pre-logit representation.}
    \label{fig:DTABL}
\end{figure}

UQ-LOB is agnostic to the choice of encoder, whose role is to map raw
LOB inputs into a useful representation. Notably, the encoder need not be
trained on the exact forecasting task: any classification task predicting
the direction of the mid-price is sufficient for pretraining. As a
lightweight encoder we adapt the Temporal Attention-Augmented Bilinear
Network (TABL) of \citet{tran2018temporal}, originally proposed for
mid-price forecasting on FI-2010. The original C(TABL) stacks two bilinear
(BL) layers followed by a single attention-augmented TABL layer, operating
on a $40\times10$ input (40 price/volume features, 10 time steps). Our data
differ from FI-2010 in feature composition and window length, so we
introduce D-TABL with two changes: (i) three BL layers rather than two,
and (ii) all layer dimensions re-derived for our $15\times512$ input (seven
continuous features plus an 8-dimensional token embedding,
Appendix~\ref{app:input_rep}, over a 512-event window). The topology, BL
layers feeding a terminal TABL layer, is unchanged (Figure~\ref{fig:DTABL}).

Each BL layer applies a bilinear transformation independently along the
feature and temporal axes,
\begin{equation}
    Y = \phi\big(W_1 X W_2 + B\big),
    \label{eq:bl}
\end{equation}
where $X\in\mathbb{R}^{D\times T}$, $W_1\in\mathbb{R}^{D'\times D}$
projects the feature axis, $W_2\in\mathbb{R}^{T\times T'}$ projects the
temporal axis, $B\in\mathbb{R}^{D'\times T'}$ is a learned bias, and
$\phi$ is a ReLU. The final layer augments Equation~\ref{eq:bl} with a
learned temporal attention mechanism~\citep{tran2018temporal}. The
feature-axis projection $\bar{X}=W_1X$ is computed first; a learned
$W\in\mathbb{R}^{T\times T}$ then produces attention scores $E=\bar{X}W$,
one per pair of time steps, normalised row-wise by a softmax,
$\alpha_{ij}=\exp(e_{ij})/\sum_k\exp(e_{ik})$, into the attention matrix
$A$. This matrix is applied element-wise to $\bar{X}$, and each time step's
representation is replaced by a convex combination of its attention-masked
and its original value,
\begin{equation}
    \tilde{X} = \lambda\big(\bar{X}\odot A\big) + (1-\lambda)\,\bar{X},
    \qquad
    Y = \phi\big(\tilde{X}W_2 + B\big),
\end{equation}
with the learned scalar $\lambda\in[0,1]$ controlling the weight of the
attended representation. The diagonal of $W$ is fixed to $1/T$ while its
off-diagonal entries are learned, so every time step has the same fixed
self-contribution and the learned entries encode the relative importance
of one time step with respect to the others.
\todo{Figure~\ref{fig:DTABL} shows a 2-way (up/down) softmax output while
Appendix~\ref{app:training_strategy} states three-class pretraining;
reconcile the figure with the text. Also state explicitly which layer's
output is used as $\mathbf{h}$ and its dimension $d_h$.}

\subsection{Model Parameter Count}
\label{app:param_count}

Table~\ref{tab:param-count} reports the parameter count of each component.
Both variants share the D-TABL encoder, the feature projection and the
context/attention trunk; the only architectural difference is the final
decoder, which produces two continuous outputs $(\mu,\nu)$ for the
regression variant and three logits for the classification variant.
\begin{table}[h]
\small
\centering
\caption{Parameter count by architectural component.}
\label{tab:param-count}
\begin{tabular}{lr}
\toprule
\textbf{Component} & \textbf{Parameters} \\
\midrule
D-TABL encoder (shared)                & 268{,}356 \\
Feature projection (shared)            & 24{,}704 \\
CNP context / attention trunk (shared) & 766{,}784 \\
Regression decoder                     & 165{,}378 \\
Classification decoder                 & 165{,}731 \\
\midrule
\textbf{Total (regression variant)}     & \textbf{1{,}225{,}222} \\
\textbf{Total (classification variant)} & \textbf{1{,}225{,}575} \\
\bottomrule
\end{tabular}
\end{table}

\subsection{Training Strategy}
\label{app:training_strategy}

Training proceeds in two stages. First, the D-TABL encoder is pretrained
on the three-class classification task alone, with three output logits
and the class-weighted cross-entropy of Section~\ref{sec:objectives}, and
no UQ head attached. A separate encoder is trained for every asset--horizon
pair on 512-event windows. Training duration per pair varies with the
asset's liquidity, since more active assets yield more windows over the
same calendar period (Table~\ref{tab:dataset-stats}); each encoder is
trained for up to 15 epochs of about 15 minutes each on a single NVIDIA
GH200 Grace Hopper Superchip, and the checkpoint with the best validation
macro F1 is retained.

Second, each UQ head is initialised from the encoder checkpoint of its
asset and horizon and fine-tuned end-to-end together with the encoder
(Section~\ref{sec:training_setup}). Both variants are trained for up to 15
epochs of about 20 minutes each on the same hardware, the additional cost
reflecting the attention-based head. The UQ-regression checkpoint with the
best validation \WR{} and the UQ-classification checkpoint with the best
validation macro F1 are retained. Across the 21 asset--horizon pairs,
base encoder (DTABL) pretraining requires about 78.75 GPU-hours and each UQ-LOB variant a
further 105 GPU-hours, for a total of about 289 GPU-hours.

\subsection{Hyperparameters}
\label{app:hyperparams}

Table~\ref{tab:hyperparams} lists all hyperparameters. Values that are
shared across assets and horizons were fixed a priori or chosen on the
validation split of BTC-USD at 5\,s and then held fixed for all other
pairs.
\todo{Fill in every entry marked ``--'' below from the training
configuration, and correct the last sentence if the selection procedure
was different.}

\begin{table}[h]
\centering
\small
\caption{Hyperparameters used for all asset--horizon pairs.}
\label{tab:hyperparams}
\resizebox{\columnwidth}{!}{%
\begin{tabular}{llr}
\toprule
\textbf{Group} & \textbf{Hyperparameter} & \textbf{Value} \\
\midrule
Input      & Window length $L$ / token vocabulary $|\mathcal{V}|$ / embedding dim. & 512 / 439 / 8 \\
           & Continuous features / max.\ depth $L_{\max}$ & 7 / 10 \\
Labels     & Latency $d$ / end-price smoothing $k$ / threshold multiplier $\alpha$ & 0.0005s / 10 / 0.25 \\
Head       & Context size $C$ / instances per batch & 15 / 16 \\
           & Projection dim.\ / attention dim.\ $d_r$ / attention heads & 128 / 256 / 4 \\
           & Context encoder / decoder depth (GELU MLP) & 3 / 5 \\
Regression & $(\lambda_1,\lambda_2,\lambda_3,\lambda_4)$ & (1.0, 0.1, 0.5, 0.5) \\
           & WMAE exponent $p$ / clamp $w_{\max}$ & 4 / 40 \\
           & Directional margin $m$ / calibration bins $B$ / variance floor $\epsilon$ & 1.0 / 3 / $1 \times 10^{-8}$ \\
Optimisation & Optimiser / learning rate (encoder, head) / weight decay & AdamW / (1e-5, 5e-5) / 0.0 \\
           & Warmup schedule & LinearLR, 10k steps \\
           & Cosine restart schedule & $T_0{=}15\text{k}$, $T_{\text{mult}}{=}2$, $\eta_{\min}{=}10^{-5}$, $\gamma{=}0.5$ \\
           & Max.\ epochs & 15 \\
           & Early-stopping metric & WR$^2$ or macro F1 \\
           & Random seed train/test & 42 / 999 \\
\bottomrule
\end{tabular}%
}
\end{table}

\subsection{Context Construction}
\label{app:context_alg}

Algorithm~\ref{alg:context} states the context construction of
Section~\ref{sec:context}. For a target window ending at $t$ with horizon
$h$ and latency $d$, the candidate context windows are those whose label
horizon has already elapsed, $e_i^{c}=t_i^{c}+d+h\le t$; the $C$ most
recent candidates form the context set. The procedure is identical at
training, validation and test time, so the test-time predictor never
conditions on information unavailable at $t$.
\todo{Confirm that the released code selects context windows exactly as
in Algorithm~\ref{alg:context} (labels realised before $t$). If the
implementation only enforces non-overlap of input windows, it must be
changed before submission, since reviewers will treat the alternative as
look-ahead.}

\begin{algorithm}[h]
\caption{Causal context construction for one target window}
\label{alg:context}
\begin{algorithmic}[1]
\REQUIRE ordered windows $\{(\mathbf{X}_j,t_j,y_j)\}$ of one asset; target index $\tau$; horizon $h$; latency $d$; context size $C$
\STATE $\mathcal{S}\leftarrow\{\,j<\tau \;:\; t_j+d+h\le t_\tau\,\}$ \hfill $\triangleright$ windows whose labels are realised by $t_\tau$
\STATE $\mathcal{C}\leftarrow$ the $C$ indices in $\mathcal{S}$ with the largest $t_j$
\RETURN context $\{(f_\theta(\mathbf{X}_j),y_j)\}_{j\in\mathcal{C}}$ and target $(f_\theta(\mathbf{X}_\tau),y_\tau)$
\end{algorithmic}
\end{algorithm}

\subsection{Input Representation}
\label{app:input_rep}

Each input window consists of 8 raw features per event: the 7 continuous
features of Section~\ref{subsec:input_representation} (log inter-event
time, log tick distance from the opposing best quote, log relative volume,
DNFI, its cumulative sums over 50 and 200 events, and QI) and one discrete
feature, the event token drawn from a vocabulary of 439 entries. A
transformer encoder consumes such tokens directly through an embedding
table, but the bilinear structure of D-TABL has no equivalent mechanism
and cannot meaningfully operate on a categorical code as if it were a
continuous quantity. We therefore embed the token before the bilinear
layers: given token id $k$ and an embedding table
$E\in\mathbb{R}^{439\times8}$, we take $e_k=E[k,:]\in\mathbb{R}^{8}$ and
concatenate it with the 7 continuous features at each time step,
\begin{equation}
    x_t = \big[\, x_t^{\text{cont}} \,;\, e_{k_t} \,\big] \in \mathbb{R}^{15},
    \qquad t = 1, \dots, 512,
\end{equation}
giving a combined input $X\in\mathbb{R}^{15\times512}$. The embedding
table is learned jointly with the rest of the network during encoder
pretraining, whose three-class label is defined in
Appendix~\ref{subsubsec:classification_labeller}.

\subsection{Labelling}
\label{app:labelling}

\subsubsection{Regression Labeller}

\textbf{Start price.} The model can only condition on information available
up to $t$, and entering a position incurs a small, fixed latency $d$ before
it can be realised. We therefore anchor the start price at the first event
at or after $t+d$: $t_{\mathrm{start}}=t+d$ and
$p_{\mathrm{start}}=p\big(\arg\min_{i:\,t_i\geq t_{\mathrm{start}}}t_i\big)$.

\textbf{End price.} The horizon $h$ is measured from $t_{\mathrm{start}}$:
$t_{\mathrm{end}}=t_{\mathrm{start}}+h$. Label statistics are computed for
$h\in\{5,10,15,30\}$\,s (Table~\ref{tab:per-horizon-thresholds}); the
experiments of the main text use $h\in\{5,10,15\}$\,s. Using a single tick
at $t_{\mathrm{end}}$ as the end price is sensitive to single-tick noise,
while averaging over the entire horizon, as in
DeepLOB~\citep{zhang2019deeplob} and
DeepFolio~\citep{sangadiev2020deepfolio}, is unsuitable at cryptocurrency
tick rates, where a horizon spans thousands of events and block-averaging
would wash out genuine displacement. We instead average the $k=10$ events
immediately at or before $t_{\mathrm{end}}$,
\begin{equation}
    p_{\mathrm{end}} = \frac{1}{k}\sum_{j \in \mathcal{K}} p_j,
    \qquad
    \mathcal{K} = \{\text{the } k \text{ events with the largest } t_j \leq t_{\mathrm{end}}\}.
    \label{eq:end_price_smooth}
\end{equation}
This never looks past $t_{\mathrm{end}}$; if fewer than $k$ events are
available the example is discarded rather than smoothed over a shorter
window.

\textbf{Target.} The regression target is the raw tick displacement
$y_h(t)=p_{\mathrm{end}}-p_{\mathrm{start}}$.

\subsubsection{Classification Labeller}
\label{subsubsec:classification_labeller}
The classification labeller reuses $t_{\mathrm{start}}$, $t_{\mathrm{end}}$,
$p_{\mathrm{start}}$ and $p_{\mathrm{end}}$ exactly as above and thresholds
the displacement $y_h(t)$ into $c\in\{\text{down},\text{up},\text{stationary}\}$.
For each asset and horizon $h$ we compute the standard deviation
$\sigma^{\mathrm{LR}}_h$ of the mid-price log-return over $h$ on the
training split and set
\begin{equation}
    \tau^{\mathrm{LR}}_h = \alpha\, \sigma_h^{\mathrm{LR}},
    \label{eq:threshold_lr}
\end{equation}
with $\alpha=0.25$ throughout. The threshold is converted from log-return
space into a tick displacement at the price level of the window,
$\delta_h(t)=p_{\mathrm{start}}\big(\exp(\tau^{\mathrm{LR}}_h)-1\big)$, and
\begin{equation}
    c =
    \begin{cases}
        1 \ (\text{up}), & y_h(t) \geq \delta_h(t) \\
        0 \ (\text{down}), & y_h(t) \leq -\delta_h(t) \\
        2 \ (\text{stationary}), & \text{otherwise.}
    \end{cases}
    \label{eq:label_rule}
\end{equation}

\subsubsection{Comparing Regression and Classification}
\label{subsubsec:compare_reg_and_cls}
To evaluate the two heads on a common footing we convert the regression
mean $\mu$ into a discrete label with the rule of
Equation~\ref{eq:label_rule}, but with a calibrated multiplier $k^{*}$ on
the threshold,
$\delta_h^{\mu}(t)=p_{\mathrm{start}}\big(\exp(k^{*}\tau^{\mathrm{LR}}_h)-1\big)$.
$k^{*}$ is selected on the held-out calibration half of the validation
period by sweeping candidate values and choosing the one whose predicted
stationary proportion most closely matches the true one on that split:
\begin{equation}
    k^{*} = \arg\min_{k} \Big| \Pr[\,|\mu| < \delta_h^{\mu,k}(t)\,] - \Pr[\,|y_h(t)| < \delta_h(t)\,] \Big|.
    \label{eq:calibration_k}
\end{equation}
Calibrating $k^{*}$ against the true class proportions, rather than the
classification head's predicted proportions, keeps the comparison
independent of any imbalance in the classifier's outputs. The threshold
$\delta_i$ in Equation~\ref{eq:dir} is $\delta_h(t)$ of the corresponding
window.

\section{Data and Label Statistics}
\label{app:data_characteristics}

\subsection{Dataset Characteristics}
\label{app:dataset-characteristics}

Table~\ref{tab:dataset-stats} summarises the event-level statistics of the
dataset of Section~\ref{sec:dataset}, computed over the train, validation
and test splits combined. Tick sizes span six orders of magnitude, from
\$0.10 for BTC to \$1$\times$10\textsuperscript{-7} for DOGE, reflecting
nominal price levels rather than market activity. Event rate (Evts/s) is
used throughout as a proxy for liquidity, and assets are sorted by it from
BTC and ETH, the most active, to LTC, the least active. This ordering does
not follow total event count or trading days, which depend on the
collection duration per asset; event rate normalises for this.

\begin{table}[t]
\centering
\scriptsize
\caption{Dataset characteristics. \textbf{Tick}: minimum price increment
(\$); \textbf{Days}: trading days; \textbf{Evts (B)}: total events
(billions); \textbf{Evts/d (M)}: events per day (millions);
\textbf{Evts/s}: events per second; \textbf{IET}: mean inter-event time (ms);
\textbf{Evts@5s/10s/15s}: expected number of events within a 5/10/15\,s
horizon (Evts/s $\times$ horizon). Assets are sorted by liquidity (Evts/s).}
\label{tab:dataset-stats}
\resizebox{\textwidth}{!}{%
\begin{tabular}{lrrrrrrrrr}
\toprule
\textbf{Sym} & \textbf{Tick} & \textbf{Days} & \textbf{Evts (B)} & \textbf{Evts/d (M)} & \textbf{Evts/s} & \textbf{IET (ms)} & \textbf{Evts@5s} & \textbf{Evts@10s} & \textbf{Evts@15s} \\
\midrule
\textbf{BTC}  & 0.10     & 94 & 1.254 & 13.338 & 156 & 6.42  & 780 & 1560 & 2340 \\
\textbf{ETH}  & 0.01     & 70 & 0.903 & 12.899 & 151 & 6.61  & 755 & 1510 & 2265 \\
\textbf{DOGE} & 1.0e{-7} & 97 & 1.124 & 11.587 & 134 & 7.46  & 670 & 1340 & 2010 \\
\textbf{SOL}  & 0.01     & 78 & 0.539 & 6.909  & 80  & 12.50 & 400 & 800  & 1200 \\
\textbf{TAO}  & 0.0001   & 84 & 0.577 & 6.870  & 80  & 12.58 & 400 & 800  & 1200 \\
\textbf{SUI}  & 0.0001   & 94 & 0.523 & 5.565  & 64  & 15.52 & 320 & 640  & 960  \\
\textbf{LTC}  & 0.01     & 83 & 0.276 & 3.326  & 39  & 25.97 & 195 & 390  & 585  \\
\bottomrule
\end{tabular}%
}
\end{table}

The last three columns translate the event rate into the expected number
of events within each horizon. This exposes a substantial disparity in
information density at a fixed horizon: at 15\,s, a BTC horizon contains
about 2{,}340 events on average against 585 for LTC, a fourfold
difference. Since the end-price smoothing window is fixed at $k=10$ events
for every asset, it covers a proportionally larger fraction of the horizon
for lower-rate assets such as LTC, SUI and TAO than for BTC, ETH or DOGE.

\begin{figure}[t]
\centering
\includegraphics[width=\textwidth]{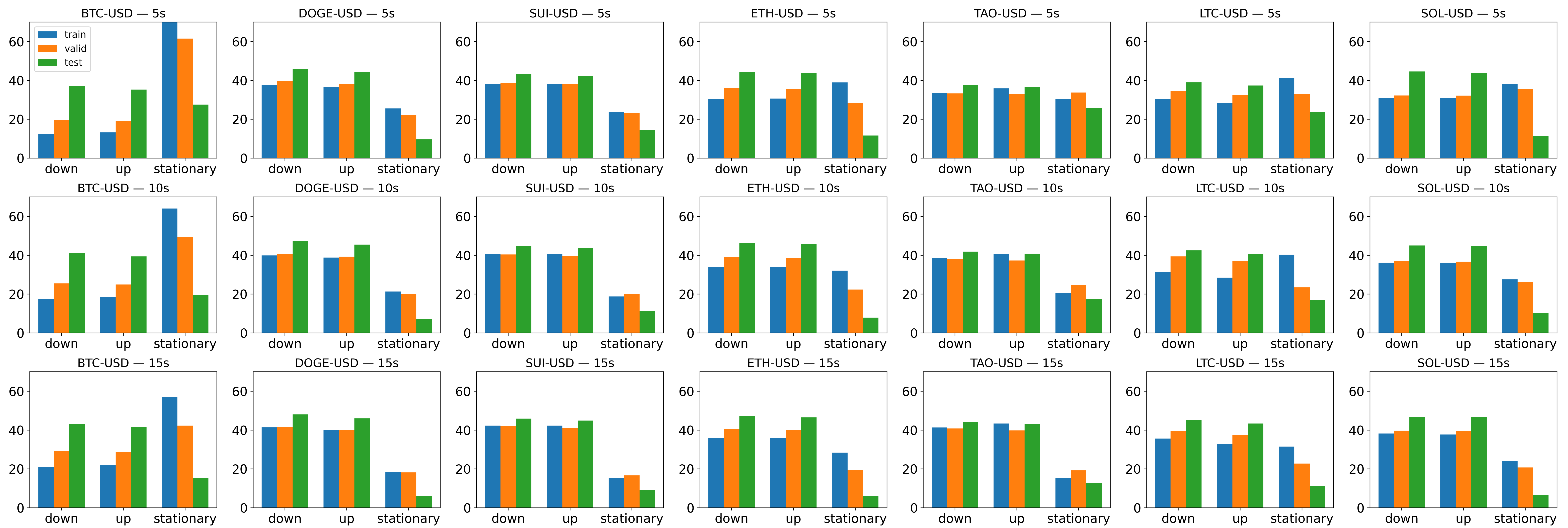}
\caption{Three-class proportions (train / validation / test) for every
asset and horizon at $\alpha=0.25$.}
\label{fig:class-proportions-grid}
\end{figure}

\subsection{Class Balance by Asset and Horizon}
\label{subsec:class-proportions}

Figure~\ref{fig:class-proportions-grid} reports the three-class
proportions for every asset and horizon at $\alpha=0.25$ across the three
splits. Two patterns hold for six of the seven assets (DOGE, SUI, ETH, TAO,
LTC, SOL): down and up proportions are closely balanced within each split,
reflecting the symmetric threshold, and the stationary proportion falls
from 5\,s to 15\,s (e.g.\ DOGE: 26\%$\to$21\%$\to$18\% in training) as
longer horizons accumulate larger movements that cross the threshold.
Train, validation and test proportions of these six assets track each
other closely at every horizon, consistent with the modest drift of
Section~\ref{subsec:class-proportion-drift}. BTC-USD is the exception: at
5\,s the training split assigns roughly 70\% of windows to the stationary
class but the test split only 27\%, the reverse of the usual pattern in
which stationary is the minority class. The gap narrows but persists at
longer horizons (train stationary 64\% at 10\,s and 57\% at 15\,s; test
19\% and 15\%).

\subsection{Class Proportion Drift}
\label{subsec:class-proportion-drift}

\begin{figure}[t]
\centering
\includegraphics[width=0.78\textwidth]{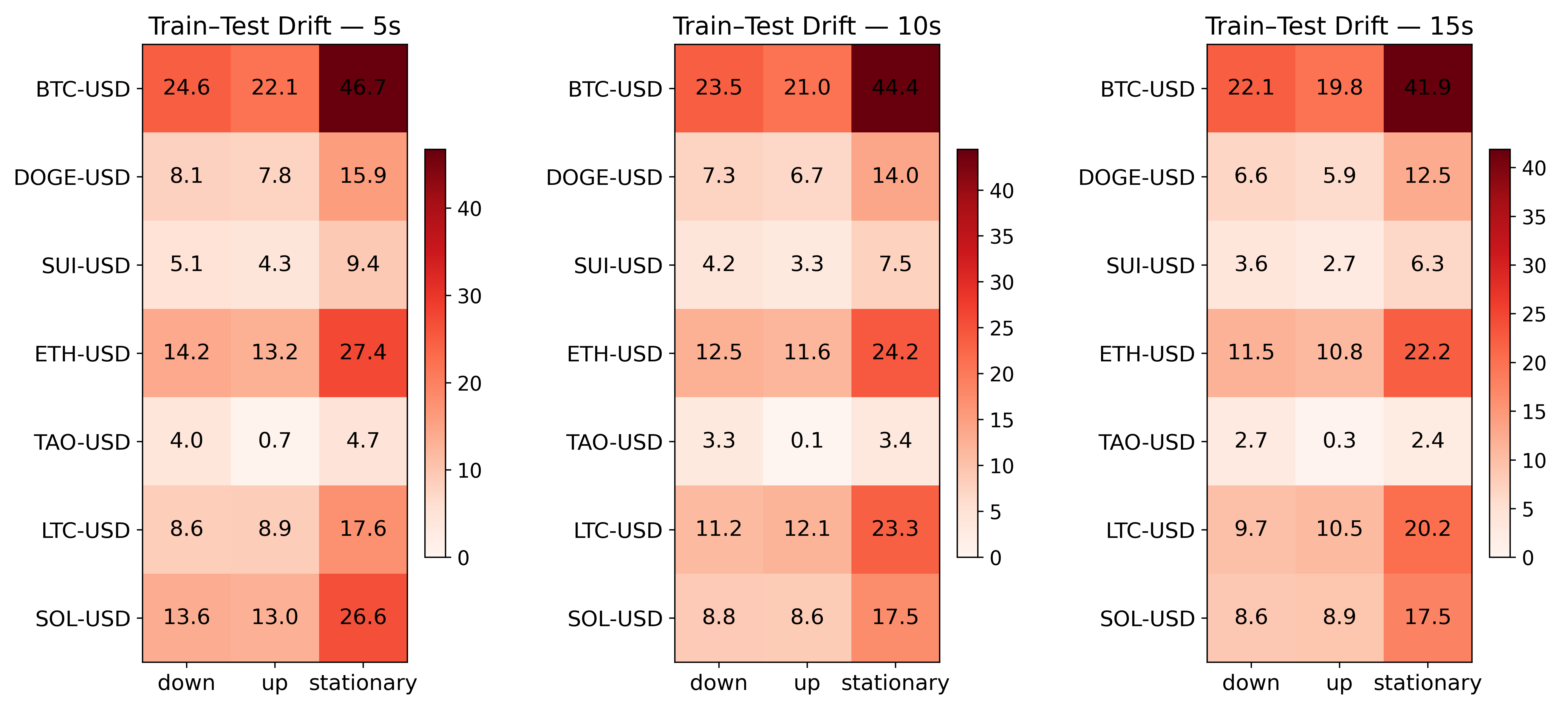}
\caption{Train--test class proportion drift $D(s,c,h)$ (percentage points)
by asset and class, across horizons, $\alpha=0.25$.}
\label{fig:class-proportion-drift}
\end{figure}

Figure~\ref{fig:class-proportion-drift} reports the absolute difference in
true class proportions between the training and test splits,
$D(s,c,h)=|P_{\text{train}}(c\mid s,h)-P_{\text{test}}(c\mid s,h)|$, for
asset $s$, class $c\in\{\text{down},\text{up},\text{stationary}\}$ and
horizon $h$. Drift varies by asset and shrinks with horizon. TAO-USD and
SUI-USD drift least, with stationary-class drift below 10 percentage
points at every horizon and 2.4\,pp and 6.3\,pp respectively at 15\,s.
ETH-USD and SOL-USD sit in a moderate range, while BTC-USD, the most liquid
asset in the dataset, drifts most, reflecting a shift in volatility regime
between the training and test windows. This shift is the main source of
distribution mismatch in our evaluation, and one motivation for
conditioning the head on recently realised outcomes.

\subsection{Per-Horizon Threshold and Scale Values}
\label{app:per-horizon-thresholds}

\begin{table}[t]
\centering
\caption{Per-asset and per-horizon log-return volatility
$\sigma^{\mathrm{LR}}_h$, label threshold $\tau^{\mathrm{LR}}_h$, and label
scale $\yref$ (ticks), computed on the training split.}
\label{tab:per-horizon-thresholds}
\resizebox{\textwidth}{!}{%
\begin{tabular}{l|rrrr|rrrr|rrrr}
\toprule
& \multicolumn{4}{c|}{$\sigma^{\mathrm{LR}}_h$} & \multicolumn{4}{c|}{$\tau^{\mathrm{LR}}_h$} & \multicolumn{4}{c}{$\yref$} \\
\cmidrule(lr){2-5}\cmidrule(lr){6-9}\cmidrule(lr){10-13}
\textbf{Symbol} & \textbf{5s} & \textbf{10s} & \textbf{15s} & \textbf{30s} & \textbf{5s} & \textbf{10s} & \textbf{15s} & \textbf{30s} & \textbf{5s} & \textbf{10s} & \textbf{15s} & \textbf{30s} \\
\midrule
\textbf{BTC}  & 1.82e-04 & 2.63e-04 & 3.29e-04 & 4.85e-04 & 4.50e-05 & 6.60e-05 & 8.20e-05 & 1.21e-04 & 155 & 181 & 205 & 255 \\
\textbf{DOGE} & 3.65e-04 & 5.12e-04 & 6.25e-04 & 8.81e-04 & 9.10e-05 & 1.28e-04 & 1.56e-04 & 2.20e-04 & 187 & 277 & 352 & 507 \\
\textbf{ETH}  & 3.14e-04 & 4.43e-04 & 5.42e-04 & 7.74e-04 & 7.80e-05 & 1.11e-04 & 1.36e-04 & 1.94e-04 &  42 &  56 &  69 &  96 \\
\textbf{LTC}  & 3.70e-04 & 5.77e-04 & 6.11e-04 & 8.94e-04 & 9.20e-05 & 1.44e-04 & 1.53e-04 & 2.23e-04 &   2 &   2 &   2 &   3 \\
\textbf{SOL}  & 3.58e-04 & 5.10e-04 & 6.18e-04 & 8.83e-04 & 8.90e-05 & 1.28e-04 & 1.55e-04 & 2.21e-04 &   3 &   3 &   4 &   6 \\
\textbf{SUI}  & 5.09e-04 & 7.34e-04 & 8.79e-04 & 1.246e-03 & 1.27e-04 & 1.83e-04 & 2.20e-04 & 3.11e-04 &   3 &   5 &   6 &   9 \\
\textbf{TAO}  & 5.01e-04 & 6.97e-04 & 8.39e-04 & 1.176e-03 & 1.25e-04 & 1.74e-04 & 2.10e-04 & 2.94e-04 & 628 & 953 & 1059 & 1596 \\
\bottomrule
\end{tabular}%
}
\end{table}

Table~\ref{tab:per-horizon-thresholds} reports, per asset and horizon, the
log-return volatility $\sigma^{\mathrm{LR}}_h$ and the resulting threshold
$\tau^{\mathrm{LR}}_h=\alpha\sigma^{\mathrm{LR}}_h$ used to construct the
three-class label (Appendix~\ref{subsubsec:classification_labeller}),
together with the label scale $\yref$ used to normalise context labels
and to scale the predicted variance in the head
(Section~\ref{sec:architecture}). All three quantities are computed
independently per asset and horizon on the training split.
\todo{State how $\yref$ is defined (e.g.\ a fixed quantile of $|y|$ on the
training split).}

\section{Extended Results}
\label{app:extended-results}

\begin{table}[t]
\centering
\caption{Per-class F1 of UQ-regression across SNR tiers, pooled over
assets, 5\,s horizon. Stationary F1 collapses under gating because
confident predictions are large in magnitude, which is why three-class
macro F1 is not a suitable gating metric.}
\label{tab:per-class-f1-snr-5s}
\scriptsize
\begin{tabular}{lrrrrr}
\toprule
\textbf{Metric} & \textbf{Top 100\%} & \textbf{Top 50\%} & \textbf{Top 30\%} & \textbf{Top 10\%} & \textbf{Top 1\%} \\
\midrule
F1 (Down)       & 0.447 & 0.526 & 0.549 & 0.598 & 0.640 \\
F1 (Up)         & 0.393 & 0.477 & 0.499 & 0.544 & 0.598 \\
F1 (Stationary) & 0.390 & 0.067 & 0.012 & 0.000 & 0.000 \\
\midrule
Macro F1        & 0.410 & 0.357 & 0.353 & 0.381 & 0.413 \\
\bottomrule
\end{tabular}
\end{table}

\begin{figure*}[h]
    \centering
    \includegraphics[width=0.95\textwidth]{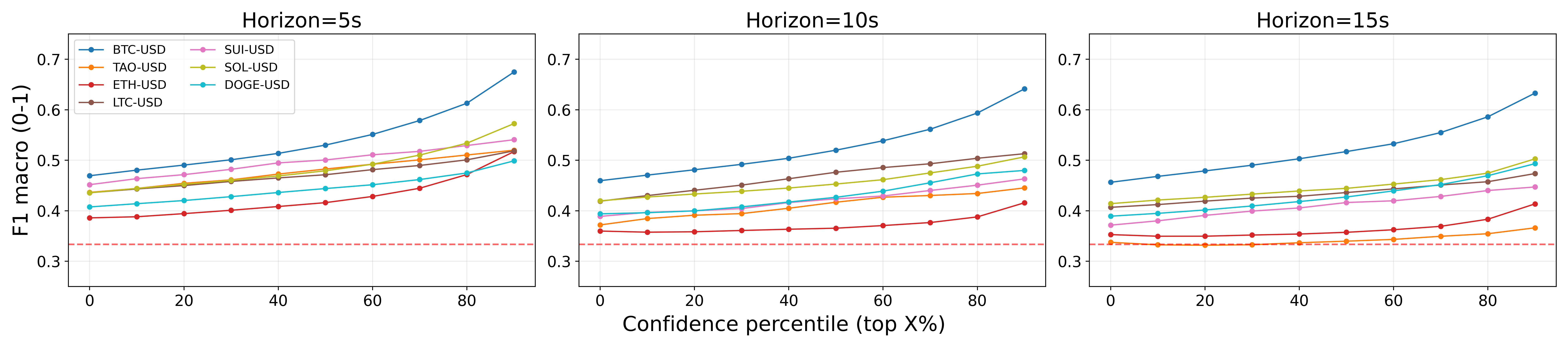}
    \caption{Three-class macro F1 of the UQ-classification head as a
    function of the softmax-confidence percentile retained, per asset and
    horizon. The dashed line is the prior-matched random reference.}
    \label{fig:softmax-vs-f1}
\end{figure*}

\begin{figure*}[h]
    \centering
    \includegraphics[width=0.95\textwidth]{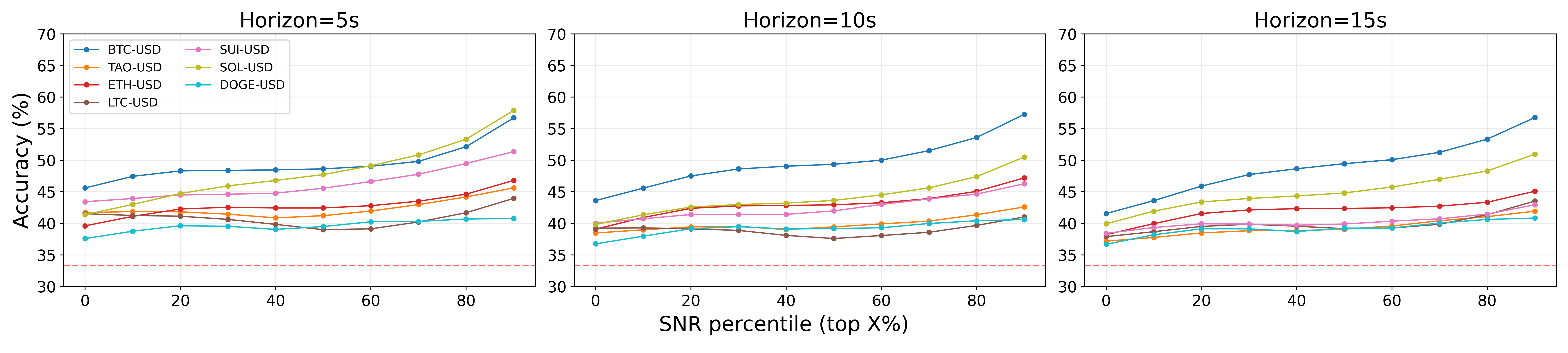}
    \vspace{0.1em}
    \includegraphics[width=0.95\textwidth]{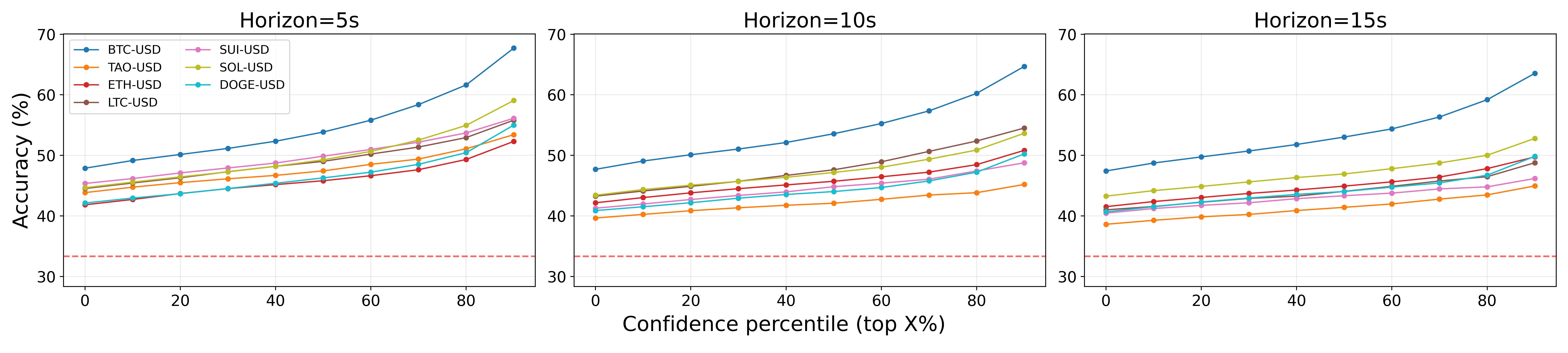}
    \caption{Three-class accuracy as a function of the confidence percentile
    retained, per asset and horizon. Top: UQ-regression gated by SNR.
    Bottom: UQ-classification gated by softmax confidence. The dashed line
    is the chance level.}
    \label{fig:classification-accuracy-results}
\end{figure*}

This appendix collects results supporting the gating analysis.
Table~\ref{tab:per-class-f1-snr-5s} decomposes three-class macro F1 by
class at the 5\,s horizon, showing the divergent behaviour of the down, up
and stationary classes under SNR gating that motivates directional macro
F1. Figure~\ref{fig:softmax-vs-f1} reports three-class macro F1 under
softmax gating for the UQ-classification head, and
Figure~\ref{fig:classification-accuracy-results} three-class accuracy
under SNR and softmax gating for the two heads, the counterparts of the
directional macro F1 curves in the main text.
\todo{Add here (i) the retained-subset prior-matched reference at each
percentile for Table~\ref{tab:directional-f1-table}, and, if available
before the deadline or for the rebuttal, (ii) per-asset tables of the main
results, (iii) multi-seed mean $\pm$ std, (iv) baselines (base D-TABL
softmax gating; MC dropout; deep ensemble) and (v) ablations (each loss
term; $C=0$ or shuffled context labels; DNFI/QI features).}

\end{document}

%% file: math_commands.tex
\usepackage{amsmath,amsfonts,bm}

\def\eqref#1{equation~\ref{#1}}

\def\1{\bm{1}}

\DeclareMathAlphabet{\mathsfit}{\encodingdefault}{\sfdefault}{m}{sl}
\SetMathAlphabet{\mathsfit}{bold}{\encodingdefault}{\sfdefault}{bx}{n}

